\documentclass{article}

\usepackage[table,dvipsnames]{xcolor}         %
\usepackage[nonatbib,preprint]{neurips_2022}

\usepackage{wrapfig}
\usepackage{tikz}
\usepackage{comment}
\usepackage{amsmath,amssymb} %
\usepackage{color}
\usepackage{graphicx}
\usepackage{amsmath}
\usepackage{amssymb}
\usepackage{booktabs}
\usepackage{tabularx}
\usepackage{pifont}
\usepackage{multirow}
\usepackage{caption}
\usepackage{subcaption}
\usepackage{booktabs}
\usepackage{lipsum}
\usepackage[export]{adjustbox}
\usepackage{longtable}
\usepackage{titletoc}
\usepackage{float}
\usepackage{xspace}
\usepackage{makecell}
\usepackage{enumitem}
\usepackage{xr}
\usepackage[percent]{overpic}
\usepackage{wrapfig}

\newcommand{\cmark}{\ding{51}}
\newcommand{\xmark}{{\color{lightgray}\ding{55}}}
\newcommand{\dash}{{\color{lightgray}--}}

\usepackage{changepage}                 %
\usepackage{lipsum}

\newlength{\offsetpage}
{\end{adjustwidth}}

\usepackage[accsupp]{axessibility}  %

\newcommand{\eg}{\emph{e.g.}\xspace}
\newcommand{\ie}{\emph{i.e.}\xspace}

\definecolor{Color4}{HTML}{FFEEEE}

\usepackage[utf8]{inputenc} %
\usepackage[T1]{fontenc}    %
\usepackage{url}            %
\usepackage{booktabs}       %
\usepackage{amsfonts}       %
\usepackage{nicefrac}       %
\definecolor{citecolor}{HTML}{2980b9}
\definecolor{linkcolor}{HTML}{c0392b}
\usepackage[pagebackref=true,breaklinks=true,colorlinks,bookmarks=false,citecolor=citecolor,linkcolor=linkcolor]{hyperref}

\definecolor{bathroomColor}{HTML}{f0b975}
\definecolor{bedroomColor}{HTML}{5ebdeb}
\definecolor{livingRoomColor}{HTML}{52ebbd}
\definecolor{kitchenColor}{HTML}{eb6a9d}

\title{AI2-THOR: An Interactive 3D Environment for Visual AI}

\author{Eric Kolve$^1$, Roozbeh Mottaghi$^{1,2}$, Winson Han$^1$, Eli VanderBilt$^1$, Luca Weihs$^1$,\\ \textbf{Alvaro Herrasti}$^1$, \textbf{Matt Deitke}$^{1,2}$, \textbf{Kiana Ehsani}$^{1}$, \textbf{Daniel Gordon}$^2$, \textbf{Yuke Zhu}$^3$,\\ \textbf{Aniruddha Kembhavi}$^{1,2}$, \textbf{Abhinav Gupta}$^{1,4}$, \textbf{Ali Farhadi}$^{1,2}$\\
$^1$Allen Institute for AI,
$^2$University of Washington,
$^3$Stanford University,
$^4$Carnegie Mellon University\\~\\
\begin{minipage}[c]{\textwidth}
\centering
    \includegraphics[width= 1\textwidth]{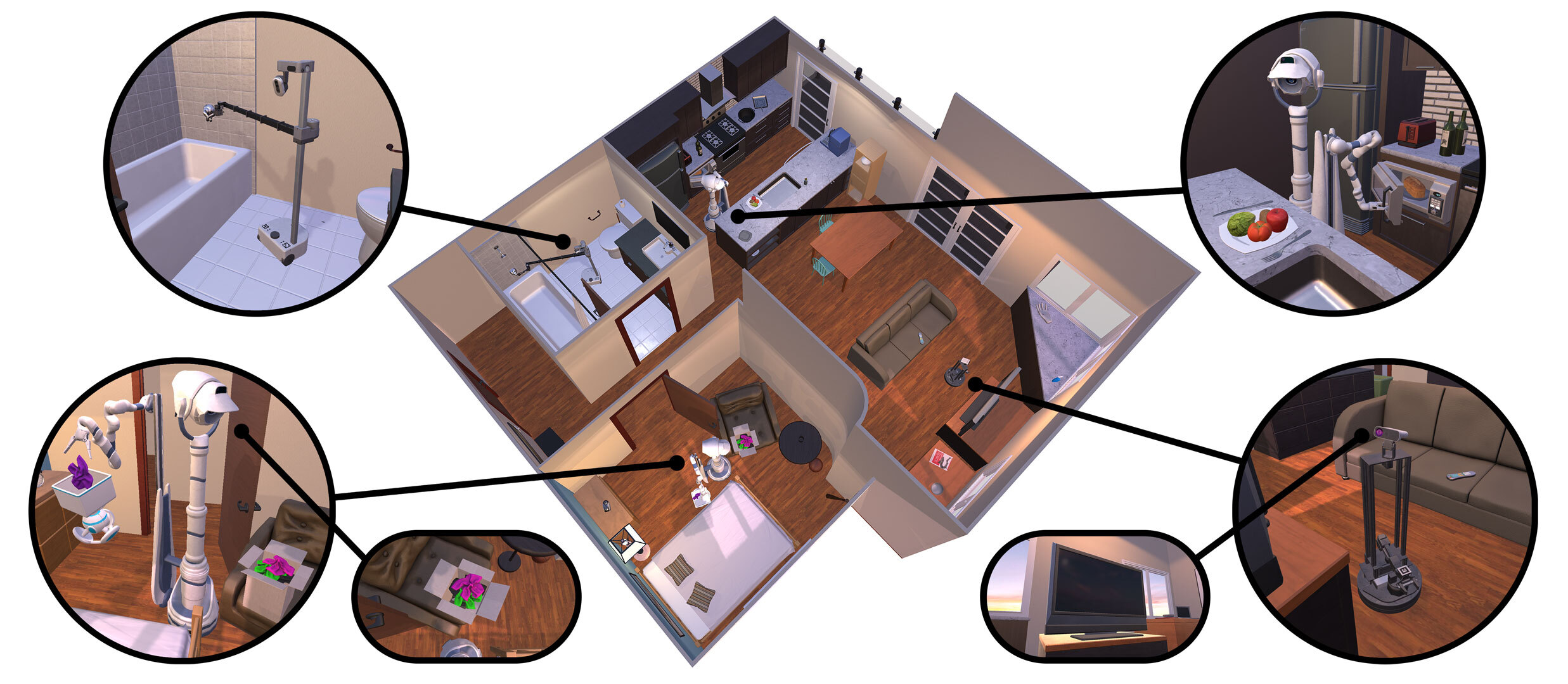}
    \captionsetup{type=figure}
    \vspace{0.025in}
\captionof{figure}{AI2-THOR consists of interactive 3D environments that can be used with embodied agents.}
\end{minipage}
}

\begin{document}

\maketitle

\begin{abstract}
We introduce The House Of inteRactions (THOR), a framework for visual AI research, available at \url{http://ai2thor.allenai.org}. AI2-THOR consists of near photo-realistic 3D indoor scenes, where AI agents can navigate in the scenes and interact with objects to perform tasks. AI2-THOR enables research in many different domains including but not limited to deep reinforcement learning, imitation learning, learning by interaction, planning, visual question answering, unsupervised representation learning, object detection and segmentation, and learning models of cognition. The goal of AI2-THOR is to facilitate building visually intelligent models and push the research forward in this domain.
\end{abstract}

\section{What is AI2-THOR?}

Humans demonstrate levels of visual understanding that go well beyond current formulations of mainstream vision tasks (\eg object detection, scene recognition, image segmentation). A key element to visual intelligence is the ability to interact with the environment and learn from those interactions. Current state-of-the-art models in computer vision are trained by using still images or videos. This is different from how humans learn. We introduce AI2-THOR as a step towards human-like learning based on visual input.

There are several key factors that distinguish AI2-THOR from other simulated environments:
\begin{enumerate}[leftmargin=0.25in]
    \item \textbf{Interactions.} AI2-THOR supports many types of interactions, including object state changes, arm-based manipulation, and causal interactions. For example, a microwave can be opened or closed, a loaf of bread can be sliced and toasted in the toaster, and a faucet can be turned on to fill a mug with water. Figure~\ref{fig:interactionFig} shows some examples of interactions supported in AI2-THOR.
    \item \textbf{Scenes.} AI2-THOR provides substantially more interactive objects and scenes for training than other platforms~\cite{li2021igibson, szot2021habitat, gan2020threedworld} by using procedural generation~\cite{deitke2022procthor}. We also provide support for many scenes designed manually by professional 3D artists, with 120 stand-alone rooms in iTHOR, 89 scenes in RoboTHOR~\cite{deitke2020robothor}, and 10 evaluation houses ArchitecTHOR~\cite{deitke2022procthor}.
    \item \textbf{Quality.} The objects and scenes in AI2-THOR are near photo-realistic. This allows better transfer of the learned models to the real world. In contrast, ATARI games or board games such as GO, which are typically used to demonstrate the performance of AI models, are very different from the real world and lack much of the visual complexity of natural environments.
    \item \textbf{API.} AI2-THOR provides a Python API to interact with the Unity 3D game engine that provides many different functionalities such as navigation, applying forces, object interaction, and physics modeling.
\end{enumerate}

Real robot experiments are typically performed in lab settings or constrained scenes since deploying robots in various indoor and outdoor scenes is not scalable. This makes training models that generalize to various situations difficult. Additionally, due to mechanical constraints of robot actuators, using learning algorithms that require thousands of iterations is infeasible. Furthermore, training real robots might be costly or unsafe as they might damage the surrounding environment or the robots themselves during training. AI2-THOR provides a scalable, fast and cheap proxy for real world experiments in different types of scenarios.

In the following sections, we discuss more of the features included in AI2-THOR, how it compares to other simulators, and work that has been conducted in it since the initial release.

\section{What does AI2-THOR feature?}

AI2-THOR is used for a wide range of tasks in Embodied AI, robotics, and computer vision. It encompasses many different types of scenes; different types of agents, each with its own set of actions to interact with objects; support for many image modalities; and functions to provide metadata about the state of the environment.

\begin{minipage}{0.4\textwidth}
\subsection{API}

Figure~\ref{fig:lifecycle} shows AI2-THOR's agent-simulator loop, which shows the front-end Python API that interacts with the Unity back-end. Here, actions are called from the Python API, which are sent through a local server to Unity. Unity is a powerful real-time game engine, which stores our scenes, code pertaining to how actions should be executed, 3D objects with their properties, and shaders to render different image modalities. Unity then returns an \texttt{Event}, which contains images from the cameras in the scene and the environment metadata.

\end{minipage}
\begin{minipage}{0.05\textwidth}\;
\end{minipage}
\begin{minipage}{0.55\textwidth}\raggedleft
\vspace{0.1in}
\centering
\includegraphics[width=\textwidth]{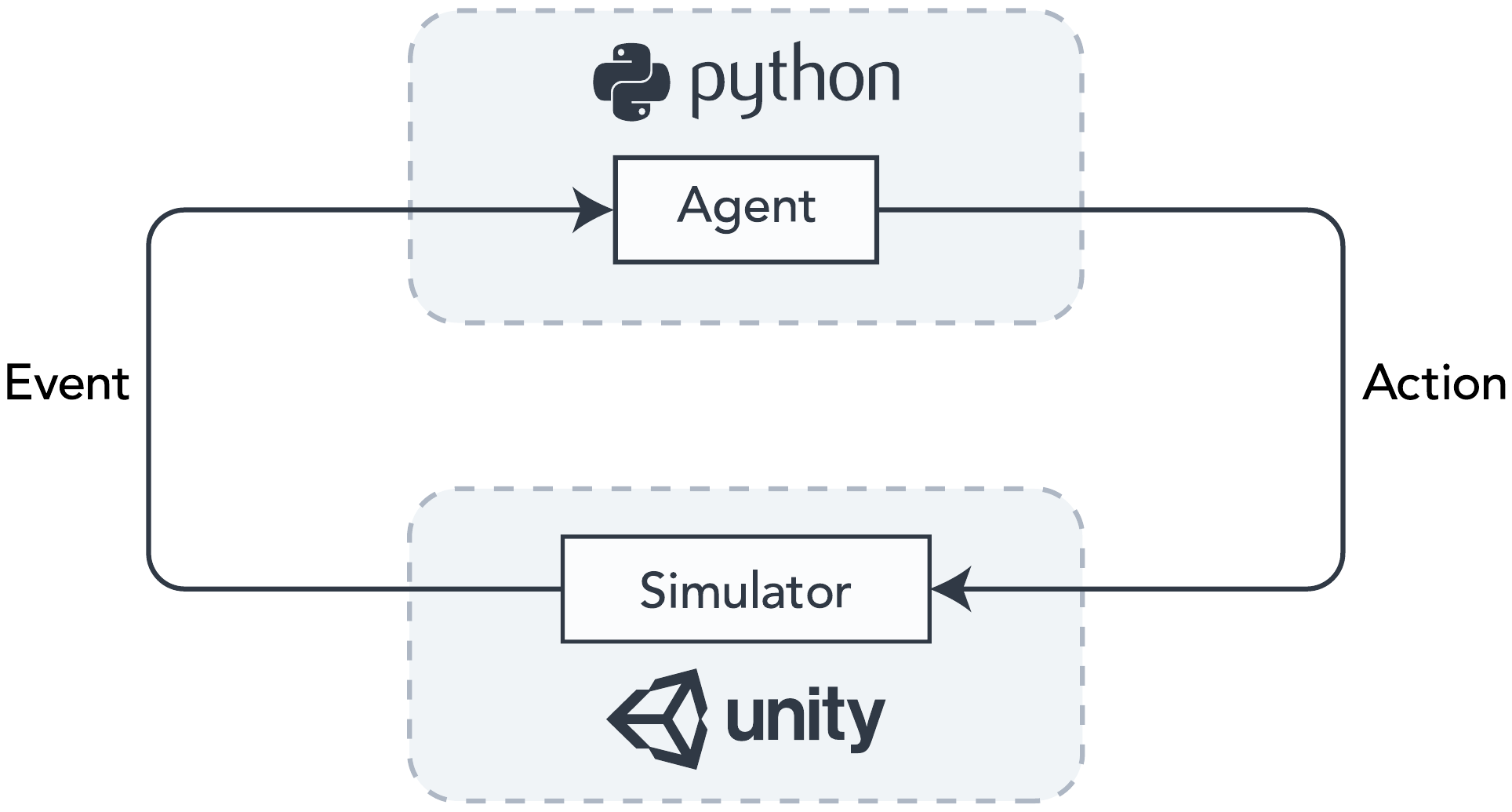}
\vspace{0in}
\captionof{figure}{AI2-THOR's agent-simulator loop, where users control an agent from the Python side that interacts with the back-end Unity simulator.}
\label{fig:lifecycle}
\end{minipage}

\subsection{Scene Datasets}

\begin{figure}[ht!]
    \centering
    \begin{subfigure}[b]{0.24\textwidth}
        \centering
        \includegraphics[width=0.6\textwidth]{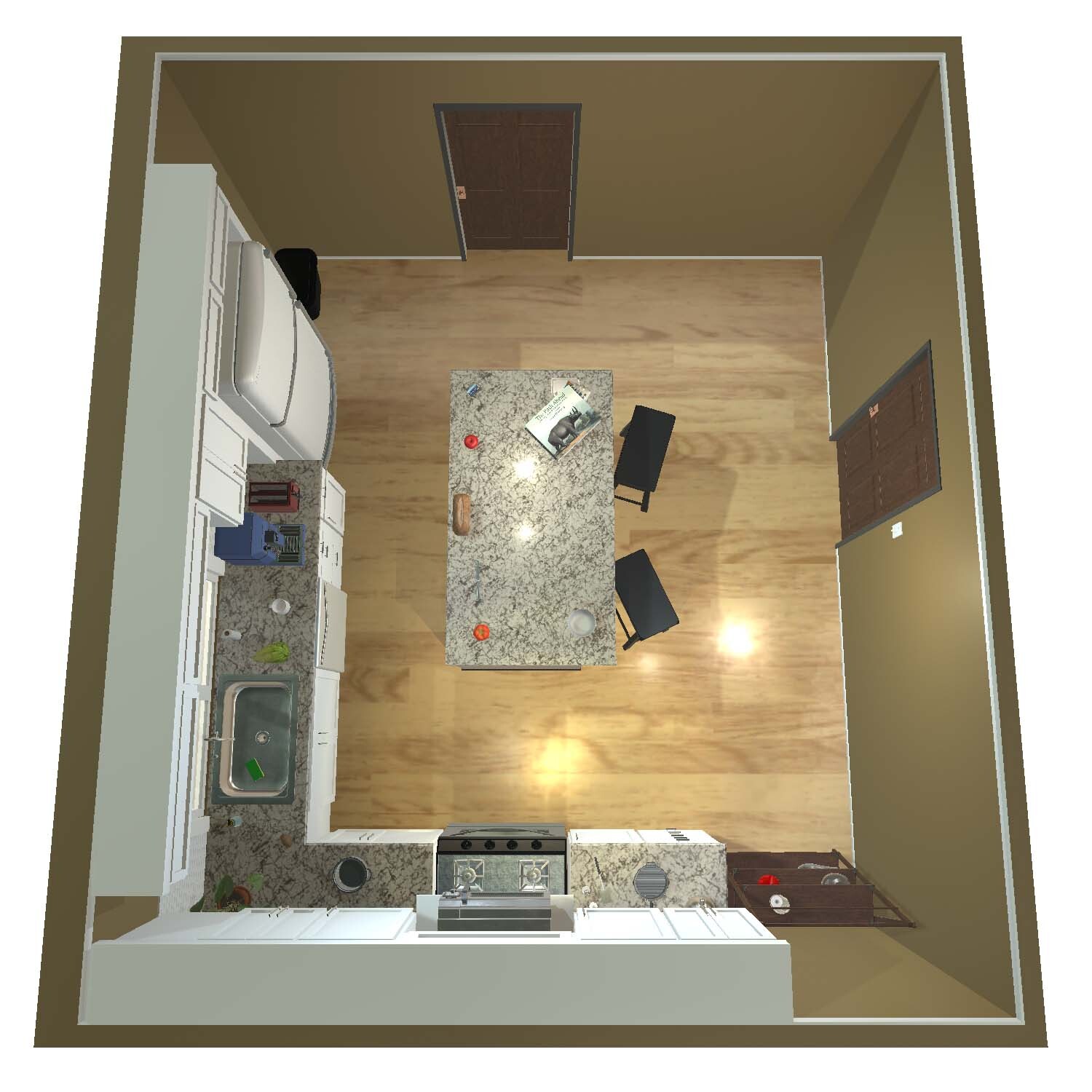}
        \vspace{0.25in}
        \caption*{iTHOR}
    \end{subfigure}
    \hfill
    \begin{subfigure}[b]{0.24\textwidth}
        \centering
        \includegraphics[width=\textwidth]{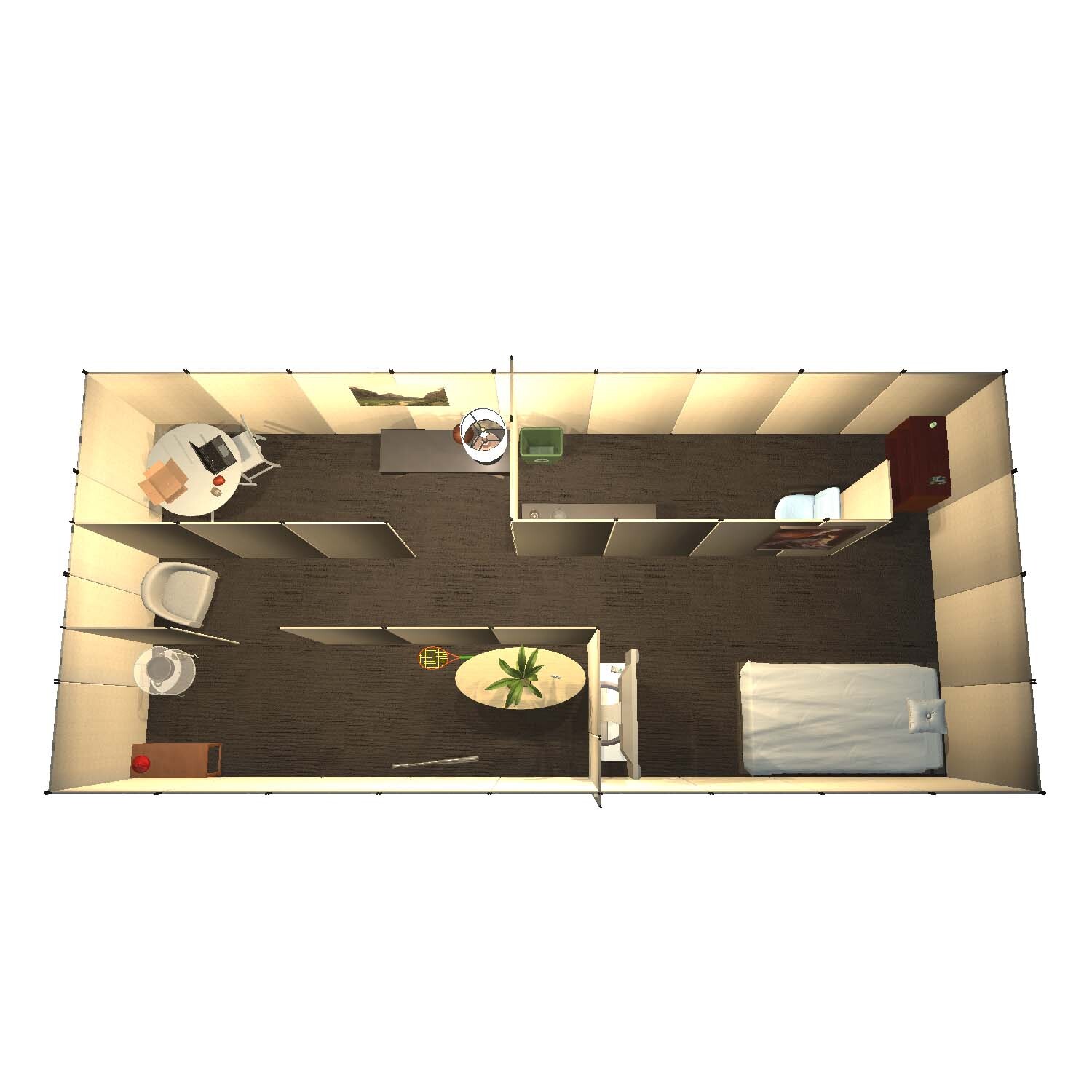}
        \caption*{RoboTHOR~\cite{deitke2020robothor}}
    \end{subfigure}
    \hfill
    \begin{subfigure}[b]{0.24\textwidth}
        \centering
        \includegraphics[width=\textwidth]{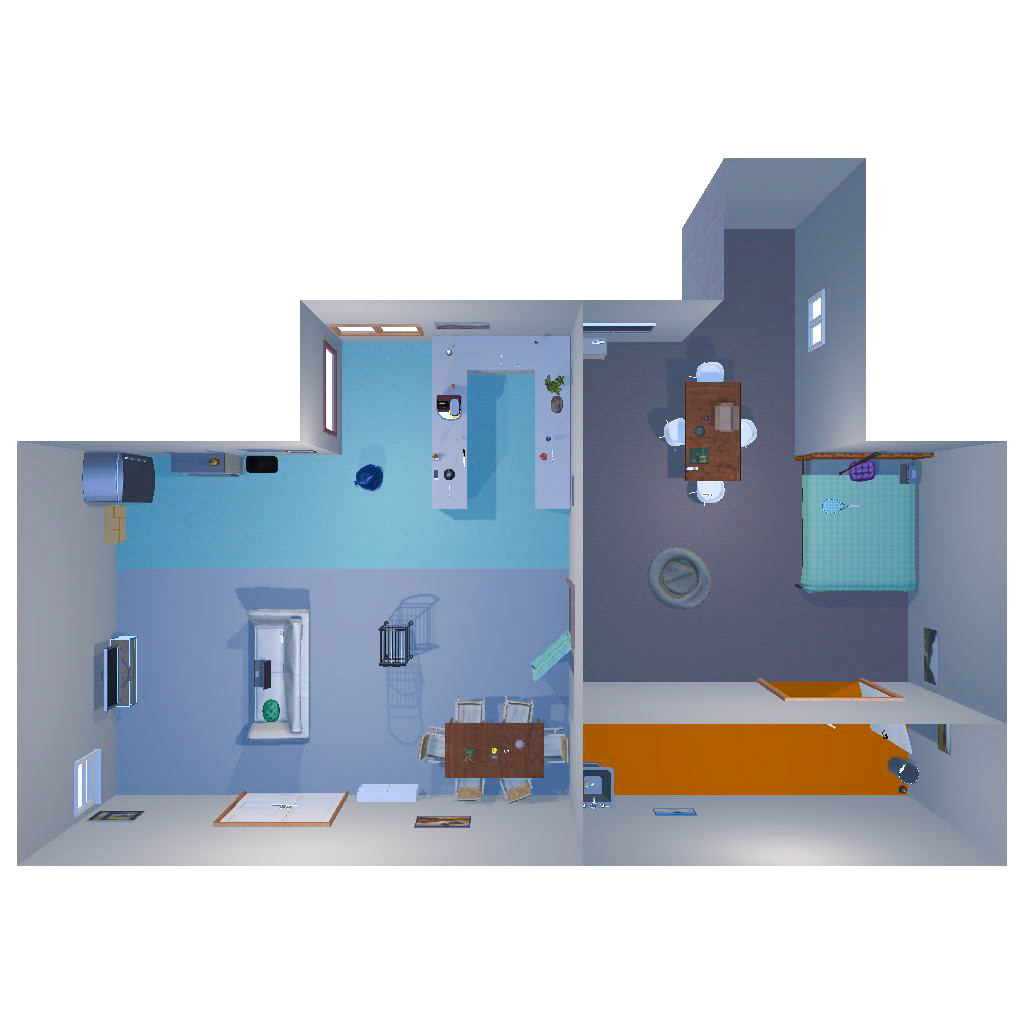}
        \caption*{ProcTHOR-10K~\cite{deitke2022procthor}}
    \end{subfigure}
    \hfill
    \begin{subfigure}[b]{0.24\textwidth}
        \centering
        \includegraphics[width=0.9\textwidth]{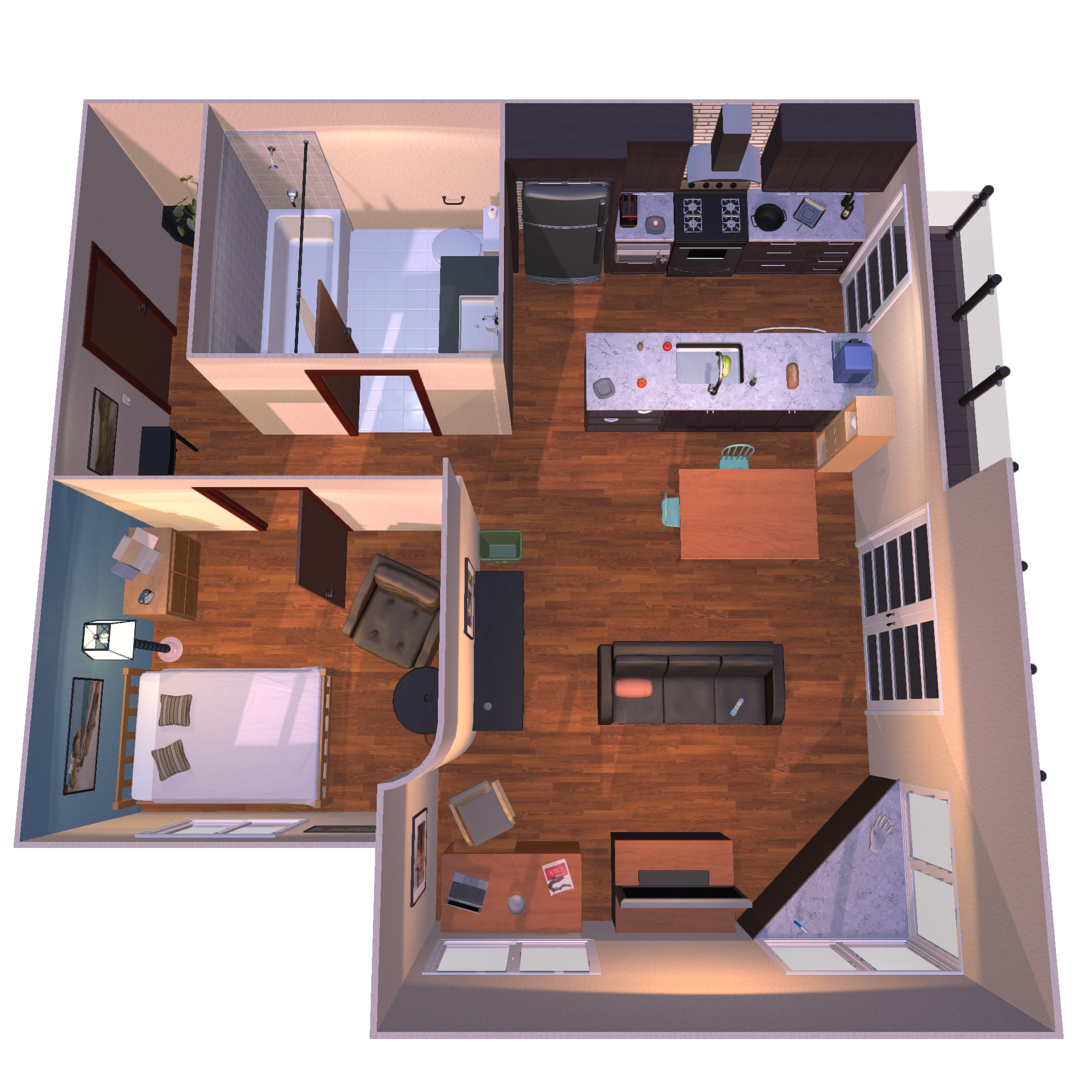}
        \vspace{0.125in}
        \caption*{ArchitecTHOR~\cite{deitke2022procthor}}
    \end{subfigure}
    \caption{AI2-THOR includes many scene datasets, including iTHOR, RoboTHOR~\cite{deitke2020robothor}, ProcTHOR~\cite{deitke2022procthor}, and ArchitecTHOR~\cite{deitke2022procthor}.}
\end{figure}

Many scene datasets have been built as part of AI2-THOR, including iTHOR, RoboTHOR~\cite{deitke2020robothor}, ProcTHOR-10K~\cite{deitke2022procthor}, and ArchitecTHOR~\cite{deitke2022procthor}. Each of these scene datasets is interactive and can be used from the same API with any of the agents.

\textbf{iTHOR} is the original set of scenes used for all experiments, which includes 120 room-sized scenes, covering bedrooms, bathrooms, kitchens, and living rooms. The scenes are modeled by hand by professional 3D artists.

\textbf{RoboTHOR}~\cite{deitke2020robothor} was later developed, which consists of 89 maze-styled dorm-sized apartments to study sim2real transfer. The scenes are also developed by professional 3D artists. Many of the scenes are recreated in Seattle, near the Allen Institute for AI's offices, to study the discrepancies when evaluating models in the same environments in simulation compared to reality.

\textbf{ProcTHOR}~\cite{deitke2022procthor} aims to use procedural generation to massively scale up the number and diversity of training scenes to improve generalization in Embodied AI. Overfitting to the training scenes is a severe problem that is often observed when training on iTHOR and RoboTHOR scenes, and it was hypothesized that merely improving the training data could help solve this problem. ProcTHOR-10K, the initial dataset released with the paper and used for experimentation, procedurally generates 10K diverse and semantically plausible houses for training. Using ProcTHOR for training led to remarkable generalization results, and we expect it to be used as a starting point for training most projects in AI2-THOR moving forward.

\textbf{ArchitecTHOR}~\cite{deitke2022procthor} is a set of 10 evaluation houses (5 for validation, 5 for testing) that was developed in conjunction with ProcTHOR. With ProcTHOR being procedurally generated, a test set of houses that comes from a real-world distribution are needed to evaluate if models training on ProcTHOR merely memorize biases from the procedural generation, or if they are capable of generalizing to real-world floorplans and object placements. Similar to iTHOR and RoboTHOR, the scenes are hand-built by professional 3D artists, although ArchitecTHOR houses are much larger and styled as single story houses.

\subsection{Agents}

\begin{figure}[ht!]
    \centering
    \begin{subfigure}[b]{0.195\textwidth}
        \centering
        \includegraphics[width=\textwidth]{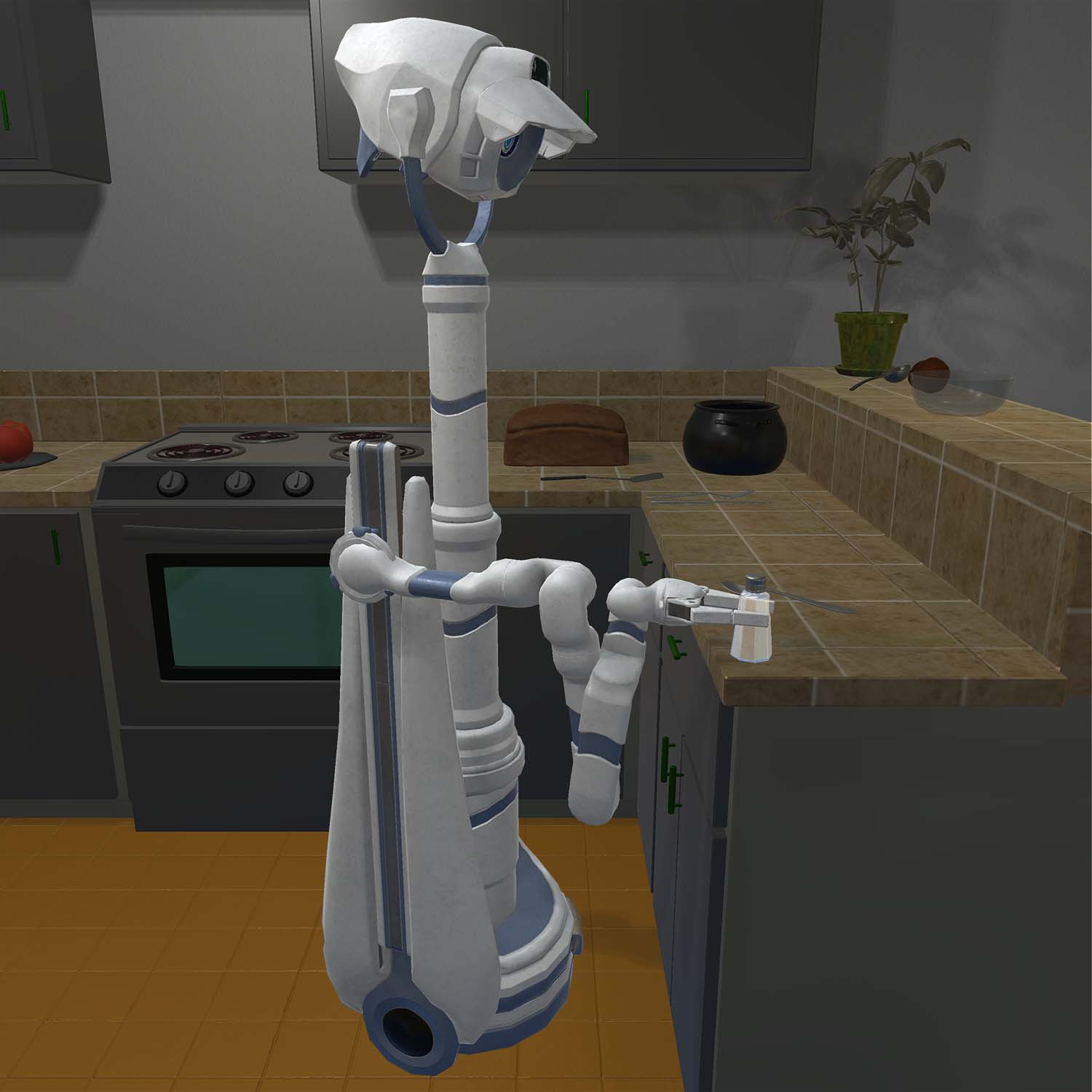}
        \caption*{ManipulaTHOR~\cite{ehsani2021manipulathor}}
    \end{subfigure}
    \begin{subfigure}[b]{0.195\textwidth}
        \centering
        \includegraphics[width=\textwidth]{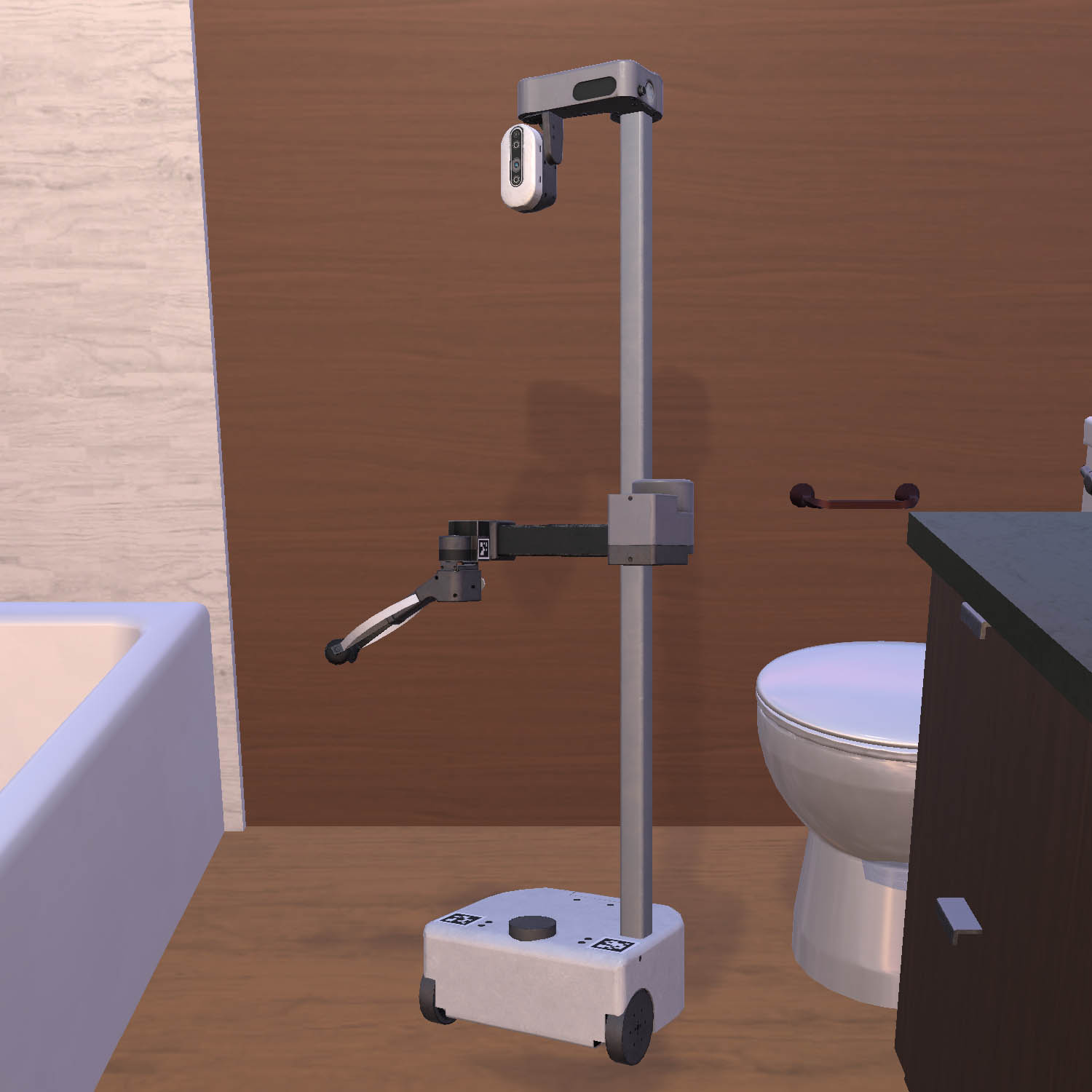}
        \caption*{StretchRE1~\cite{kemp2022design}}
    \end{subfigure}
    \begin{subfigure}[b]{0.195\textwidth}
        \centering
        \includegraphics[width=\textwidth]{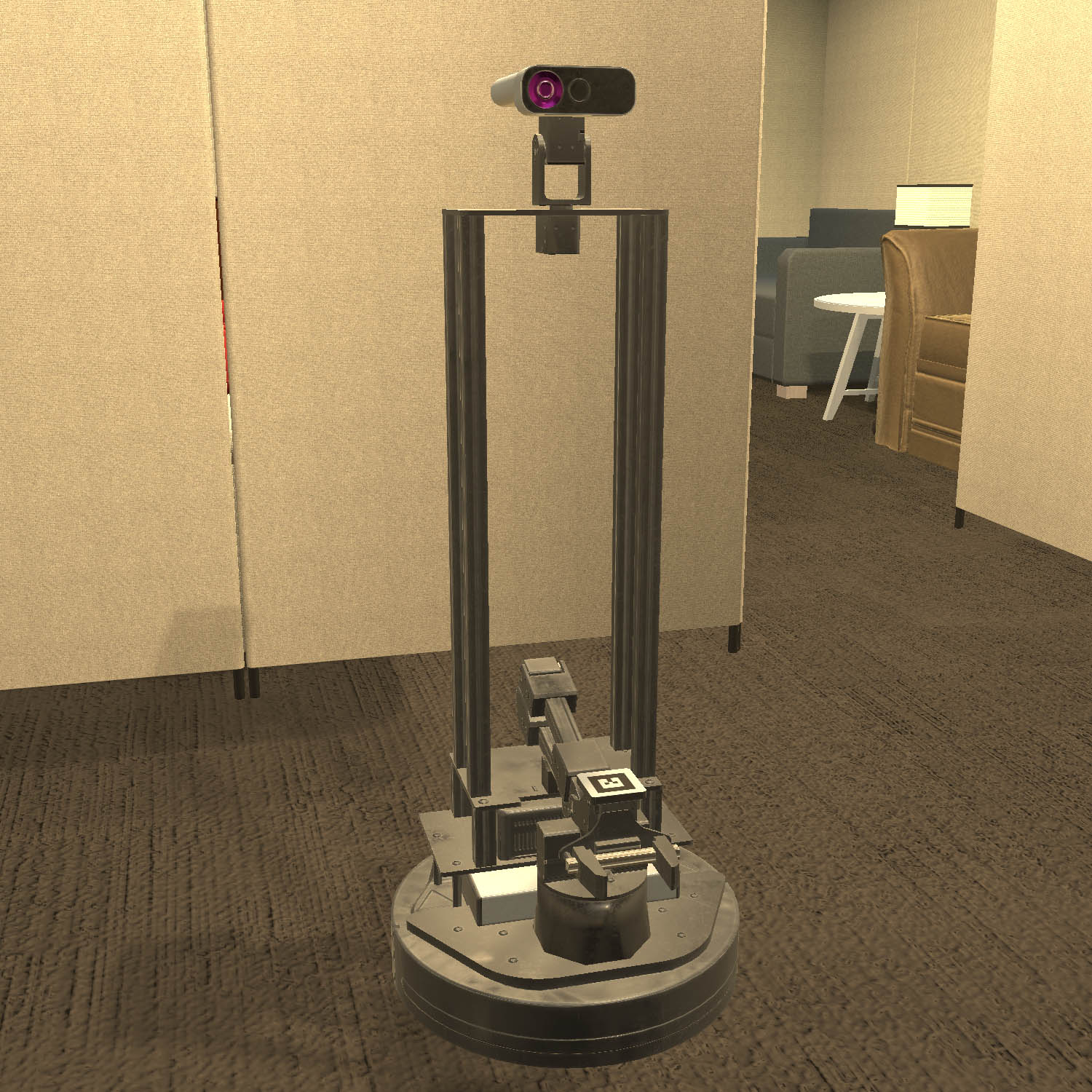}
        \caption*{LoCoBot~\cite{murali2019pyrobot}}
    \end{subfigure}
    \begin{subfigure}[b]{0.195\textwidth}
        \centering
        \includegraphics[width=\textwidth]{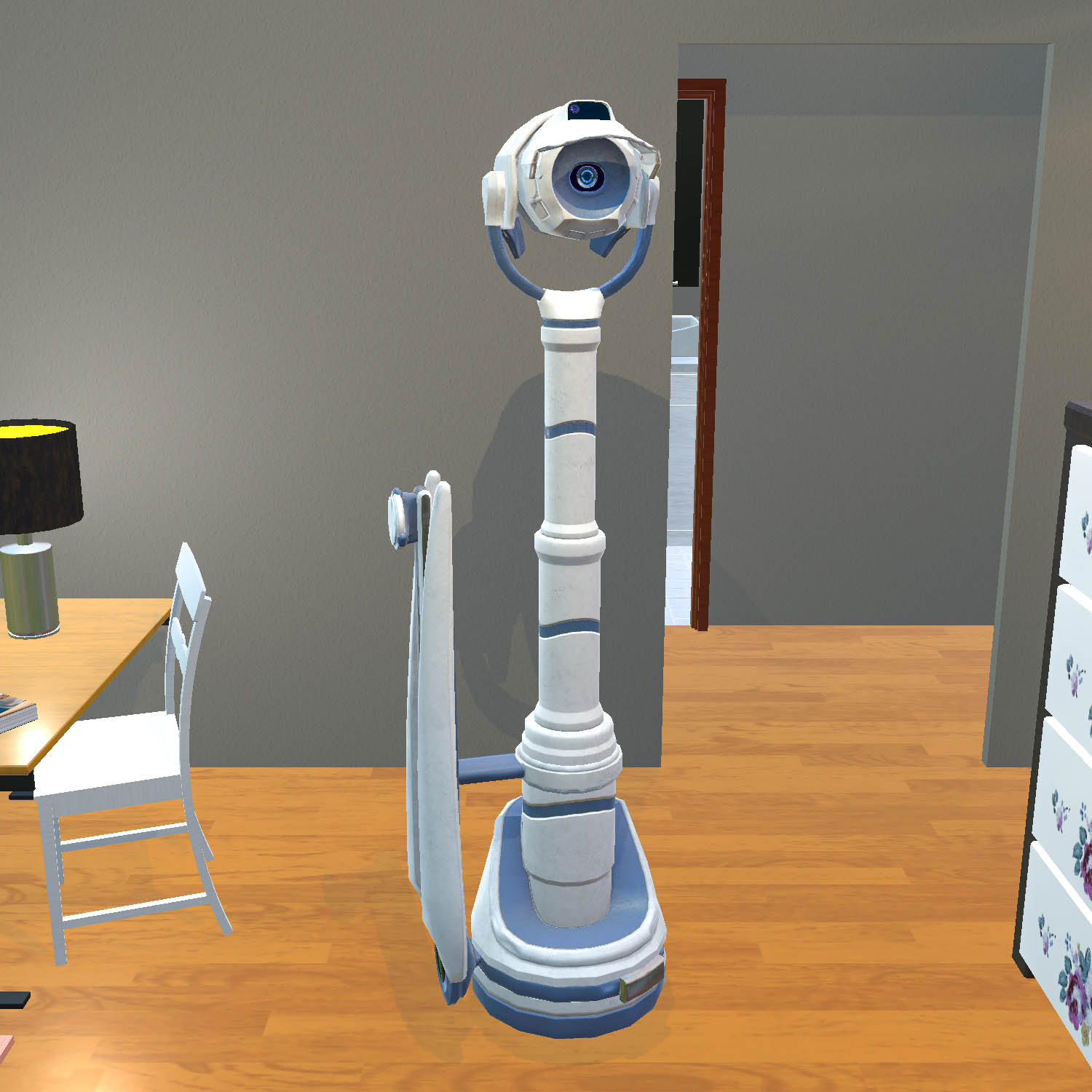}
        \caption*{Abstract}
    \end{subfigure}
    \begin{subfigure}[b]{0.195\textwidth}
        \centering
        \includegraphics[width=\textwidth]{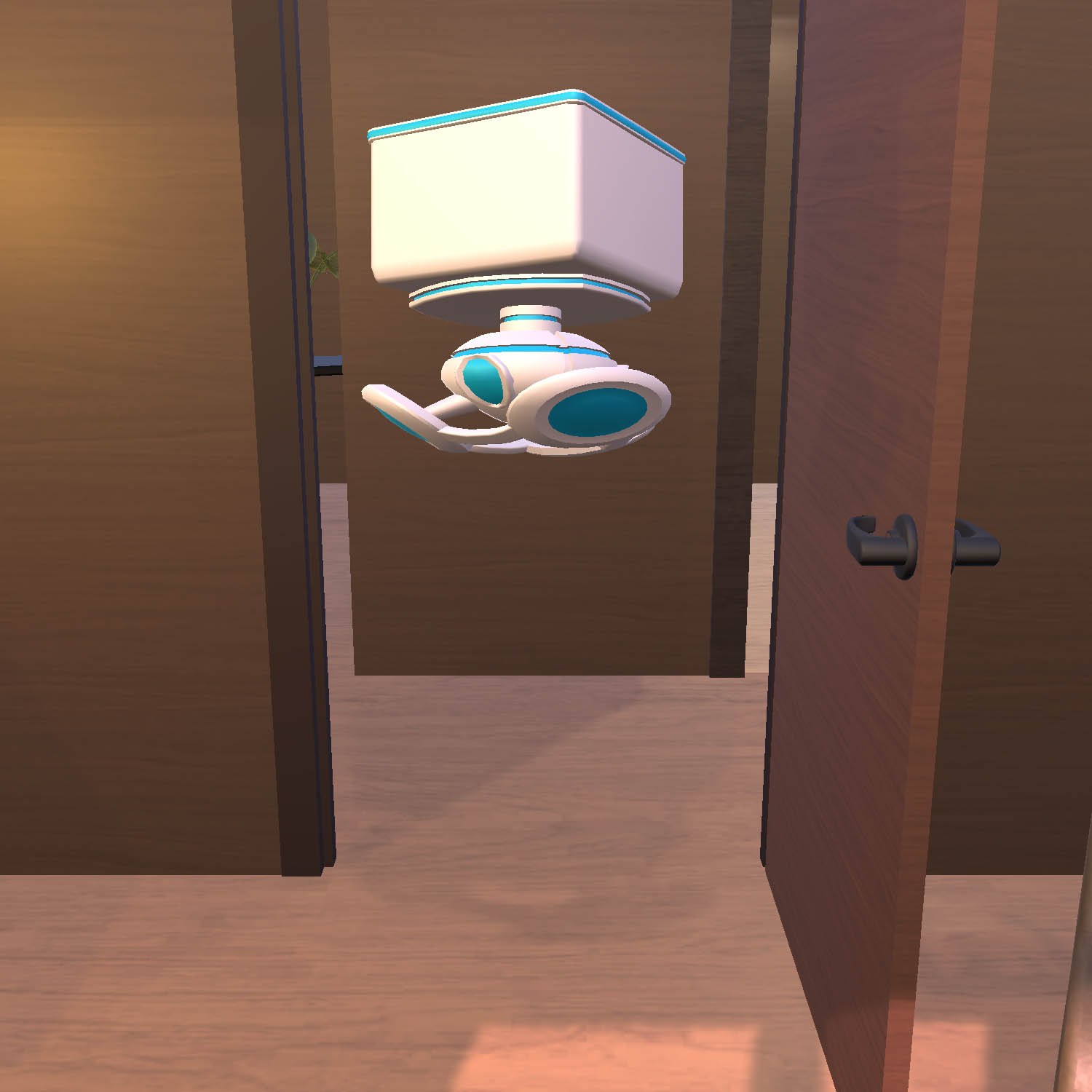}
        \caption*{Drone~\cite{zeng2020visual}}
    \end{subfigure}
    \caption{The current agents available in AI2-THOR include the ManipulaTHOR and StretchRE1 agents, which support arm manipulation, and the LoCoBot, Abstract, and Drone agents, which support navigation and abstracted interaction.}
\end{figure}

AI2-THOR comes equipped with many agents that support a range of embodiments, including the ManipulaTHOR~\cite{ehsani2021manipulathor} agent, StretchRE1~\cite{kemp2022design}, LoCoBot~\cite{murali2019pyrobot}, Abstract agent, and Drone~\cite{zeng2020visual} agent. Each of these agents is embodied with a different physical robot and has its own set of actions that it can execute in the environment.

All of the agents are able to navigate around the scenes and perform environment queries and state changes. The ManipulaTHOR and StetchRE1 agents are able to use their arm to grasp and open objects. The LoCoBot, Abstract, and Drone agents interact with objects in a more abstract way, where a high-level \textsc{Open} or \textsc{Pickup} command is executed if the agent is looking at the object, the high-level action is called, and the agent within a certain distance of the object.

\subsection{Actions}

Agents in AI2-THOR support a wide range of actions, which we can break down into navigation actions, interaction actions, environment queries, and environment state changes.

\begin{minipage}[c]{\textwidth}
    \centering
    \captionsetup{type=figure}
    \begin{subfigure}[b]{0.32\textwidth}
        \captionsetup{type=figure}
        \includegraphics[width=\textwidth]{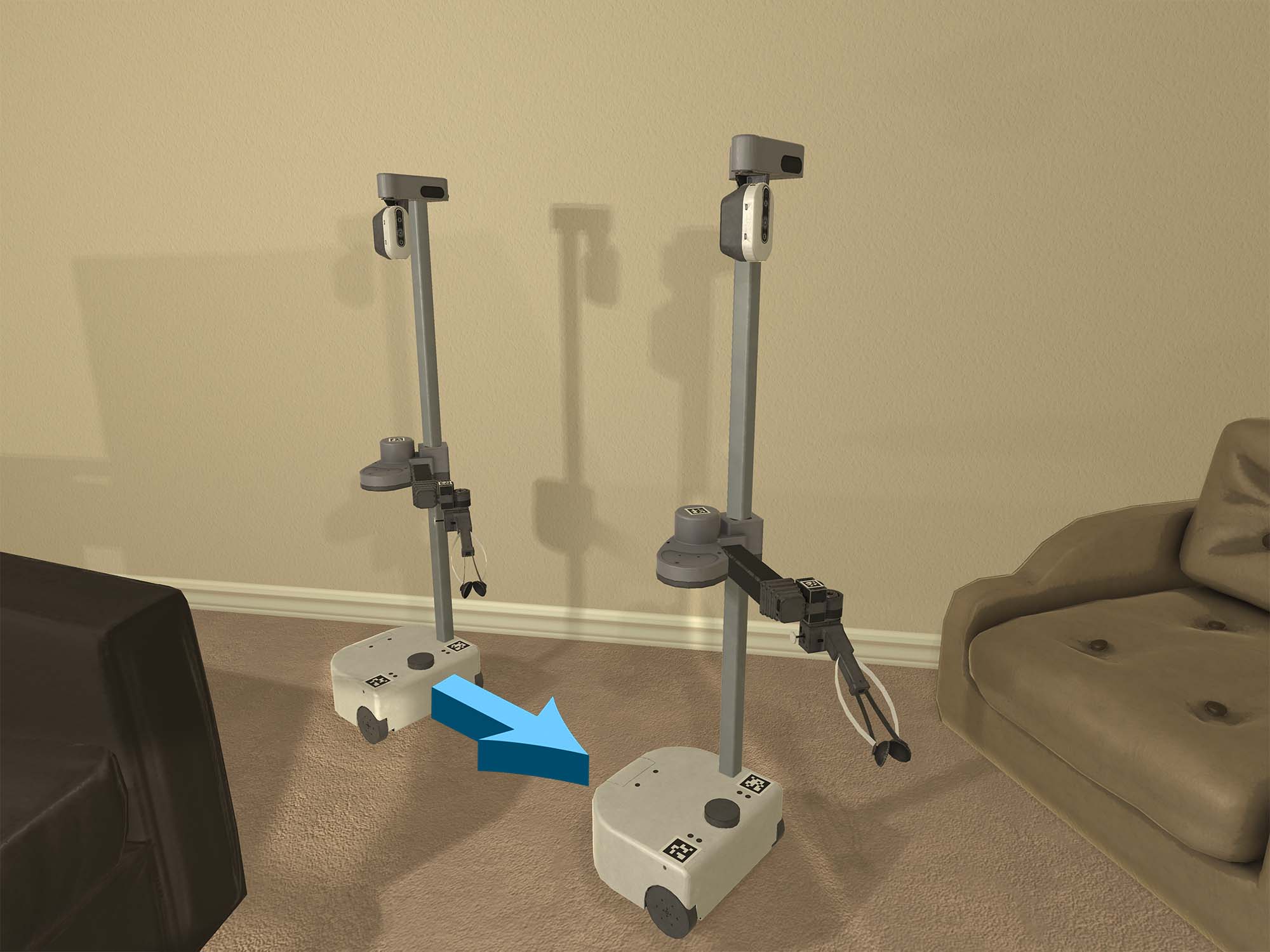}
        \captionof{figure}{{\color{Blue}Navigating}}
    \end{subfigure}
    \hfill
    \begin{subfigure}[b]{0.32\textwidth}
        \captionsetup{type=figure}
        \includegraphics[width=\textwidth]{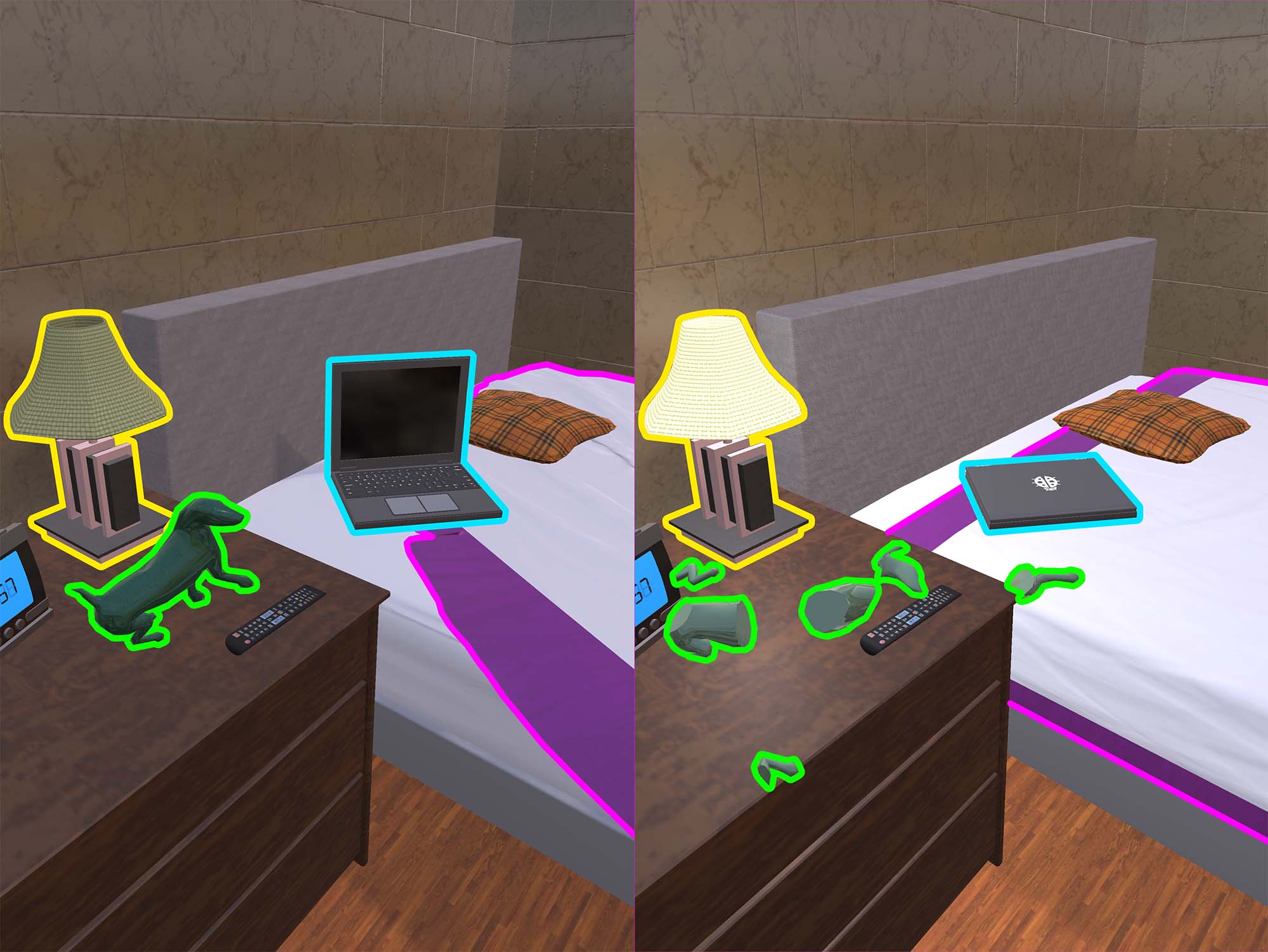}
        \captionof{figure}{{\color{Maroon}Changing Object States}}
    \end{subfigure}
    \hfill
    \begin{subfigure}[b]{0.32\textwidth}
        \captionsetup{type=figure}
        \includegraphics[width=\textwidth]{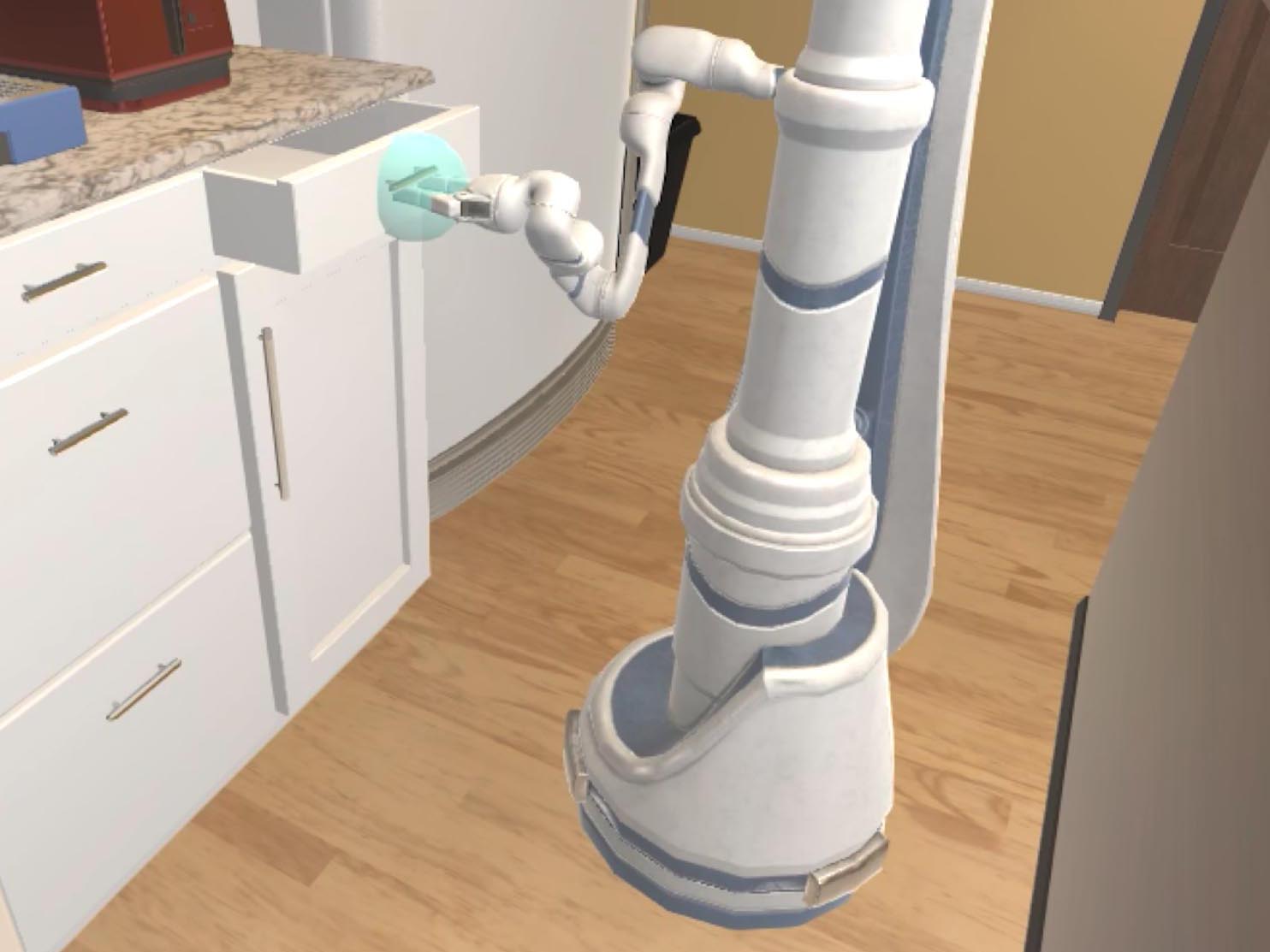}
        \captionof{figure}{{\color{Maroon}Opening an Object}}
        \label{fig:open}
    \end{subfigure}
    \hfill
    \vspace{0.1in}
    \begin{subfigure}[b]{0.32\textwidth}
        \captionsetup{type=figure}
        \includegraphics[width=\textwidth]{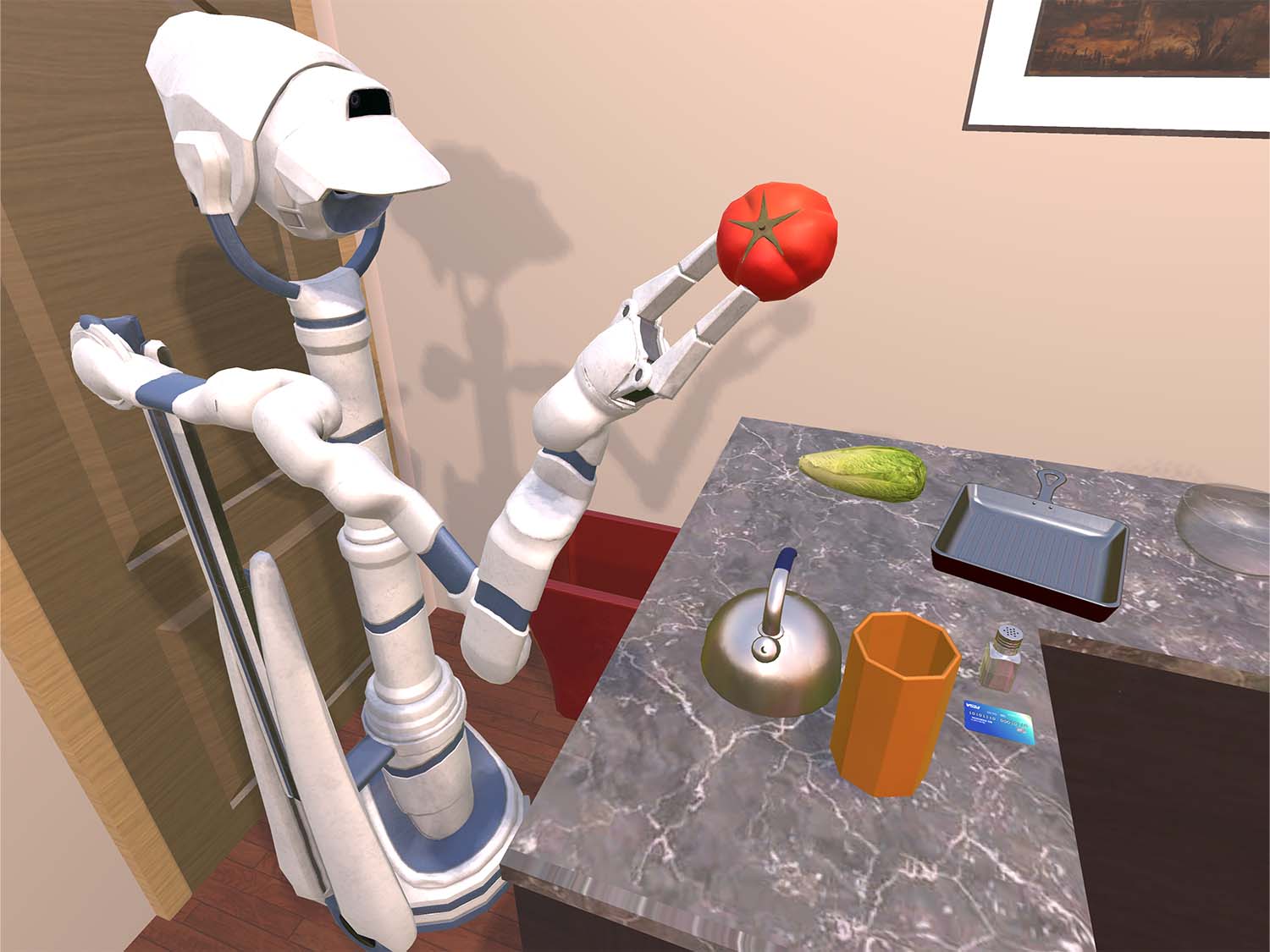}
        \captionof{figure}{{\color{Maroon}Grasping an Object}}
        \label{fig:grasp}
    \end{subfigure}
    \hfill
    \begin{subfigure}[b]{0.32\textwidth}
        \captionsetup{type=figure}
        \includegraphics[width=\textwidth]{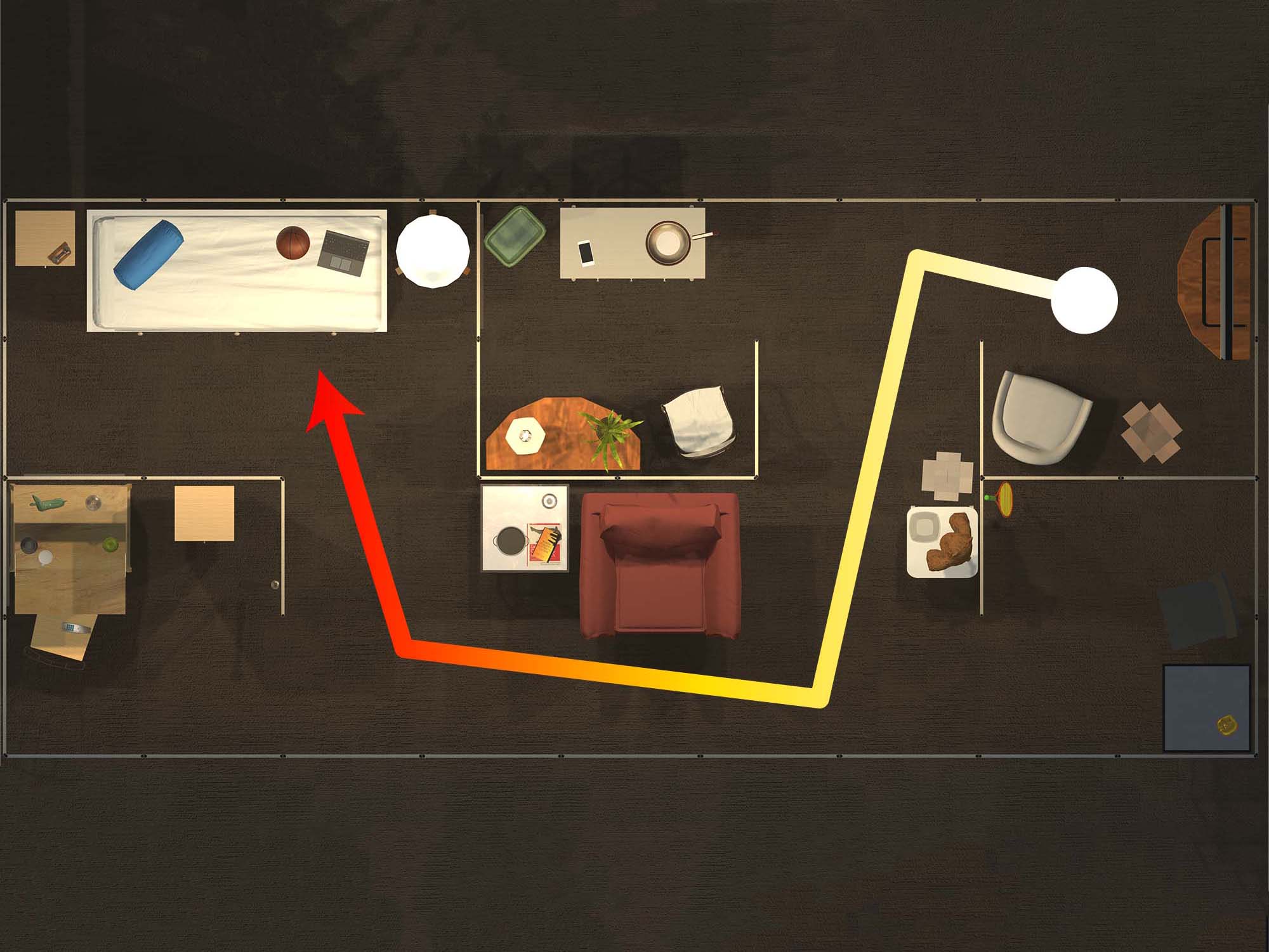}
        \captionof{figure}{{\color{SeaGreen}Finding the Shortest Path}}
    \end{subfigure}
    \hfill
    \begin{subfigure}[b]{0.32\textwidth}
        \captionsetup{type=figure}
        \includegraphics[width=\textwidth]{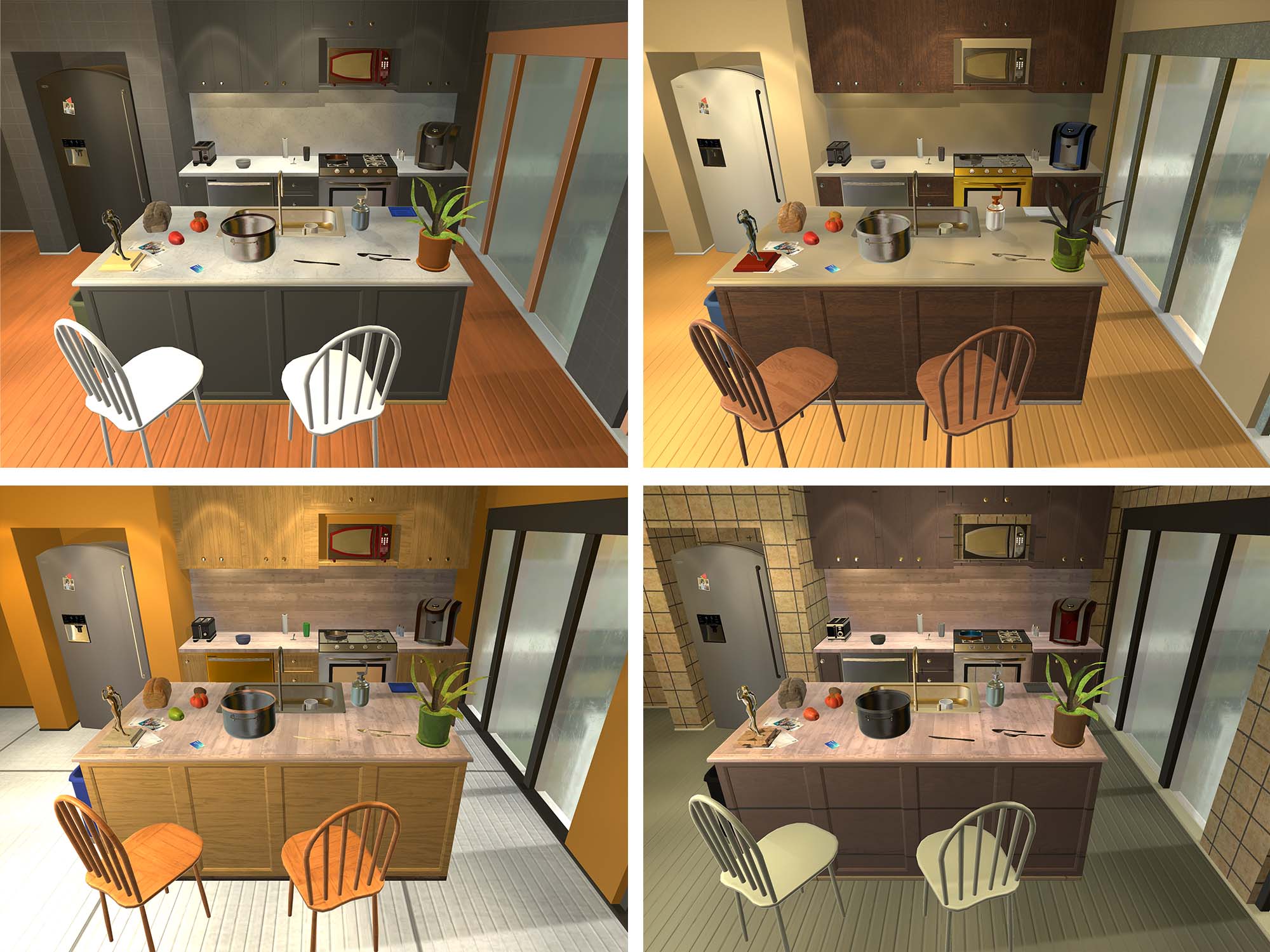}
        \captionof{figure}{{\color{YellowOrange}Randomizing Materials}}
        \label{fig:randmat}
    \end{subfigure}
    \captionof{figure}{Examples of actions supported in AI2-THOR, including {\color{Blue}navigation actions} (\eg movement), {\color{Maroon}interactive actions} (\eg object state changes and grasping), {\color{SeaGreen}environment queries} (\eg finding the shortest path), and {\color{YellowOrange}environment state changes} (\eg randomizing materials).}
    \label{fig:interactionFig}
\end{minipage}

\paragraph{Navigation Actions.} Each agent comes with some ability to navigate in a given scene. Navigation actions may be discrete or continuous move (\eg \textsc{MoveAhead} by 0.25m), rotate (\eg \textsc{RotateRight} by 30$^\circ$), look (\eg \textsc{LookUp} by 30$^\circ$), or teleport actions. Agents with an arm have more actions to control how the arm is positioned.

\paragraph{Interactive Actions.} There are many types of interactions supported in AI2-THOR, including abstracted interactions, arm-based manipulation, object state changes, and causal interactions.

Abstracted interactions are often a key component of research in Embodied AI, where one may be interested in studying high-level planning rather than low-level control. Here, an agent can execute an abstracted action, such as open, pickup, push, throw, drop, or place, where as long as the agent can see the object in its frame and it is within a certain distance away from it, the action can execute successfully. Abstracted actions can be used to change an object's state, such as cooking it, breaking it, slicing it, toggling it, filling it with liquid, or using it up.

Arm-based interactions are lower-level than abstracted actions, and require interacting with objects by moving an arm to grip them. They can be used to open an object incrementally in a continuous manner (Figure~\ref{fig:open}) or grasping an object to move it from one position to another (Figure~\ref{fig:grasp}).

Causal interactions result as a consequence of interacting with another object. For instance, turning a coffee machine on, which has a mug placed in it, will fill the mug with coffee; throwing a breakable an object hard enough may cause it and the surface it is thrown at to shatter; and pushing a table over will cause objects on top of the table to fall and potentially break.

\paragraph{Environment Queries.} Environment queries are used to obtain information about the state of the environment that is not provided with each \textsc{Event} because it is often unnecessary to compute at each time step for every use case. Examples include obtaining the shortest path from the agent to a given target object in the scene, querying which object appears at a pixel in the agent's current frame, or obtaining the convex hull of a given object.

\paragraph{Environment State Changes.} Environment state changes involve actions that modify the environment or its properties. For example, some environment state changes include randomizing the materials in the scene (Figure~\ref{fig:randmat}), randomizing the lighting in the scene, updating the rendering quality, updating the resolution of the images from the cameras, and changing the skybox in the scene.

\subsection{Image Modalities}

\begin{figure}[ht!]
    \centering
    \begin{subfigure}[b]{0.195\textwidth}
         \centering
         \includegraphics[width=\textwidth]{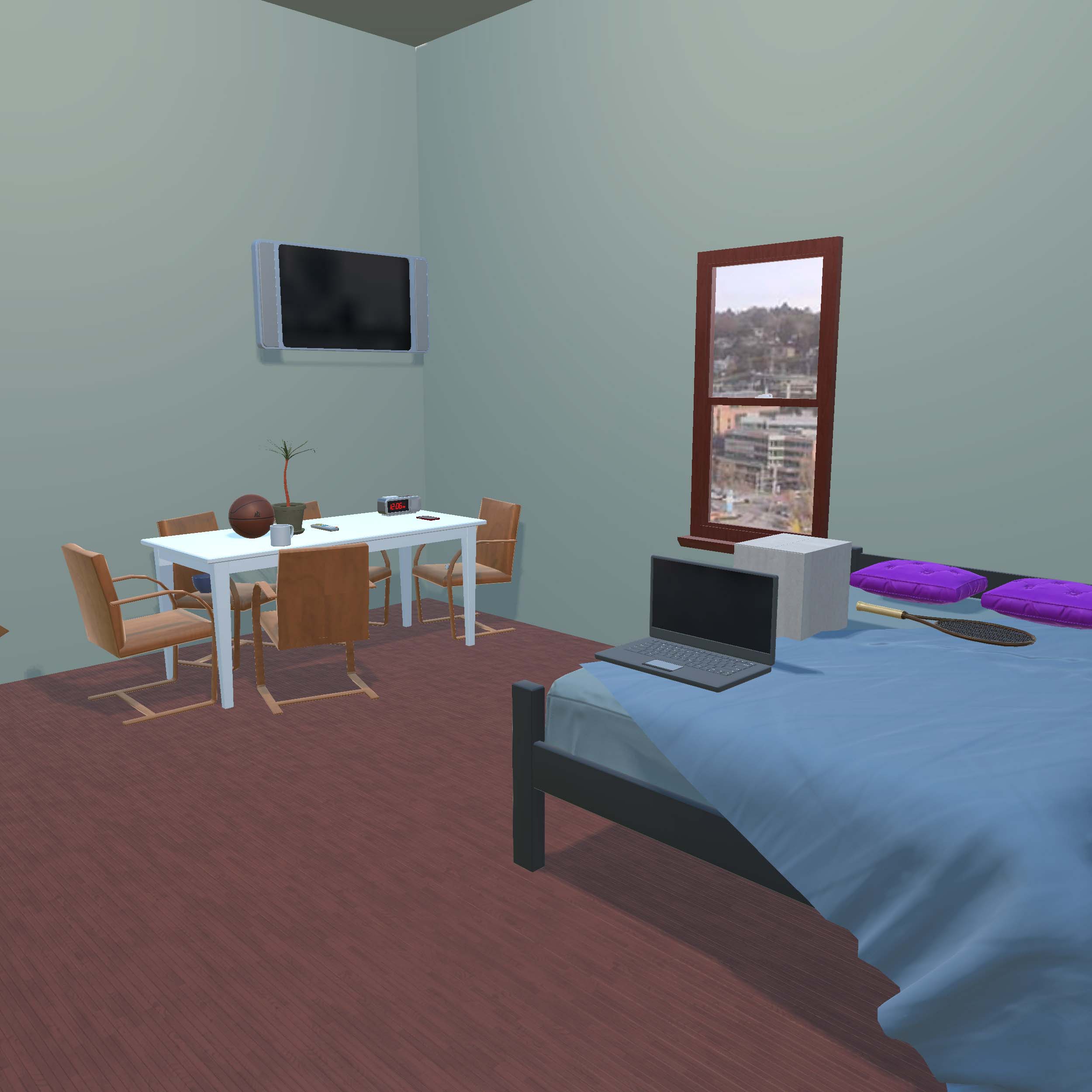}
         \caption*{RGB}
     \end{subfigure}
     \hfill
     \begin{subfigure}[b]{0.195\textwidth}
         \centering
         \includegraphics[width=\textwidth]{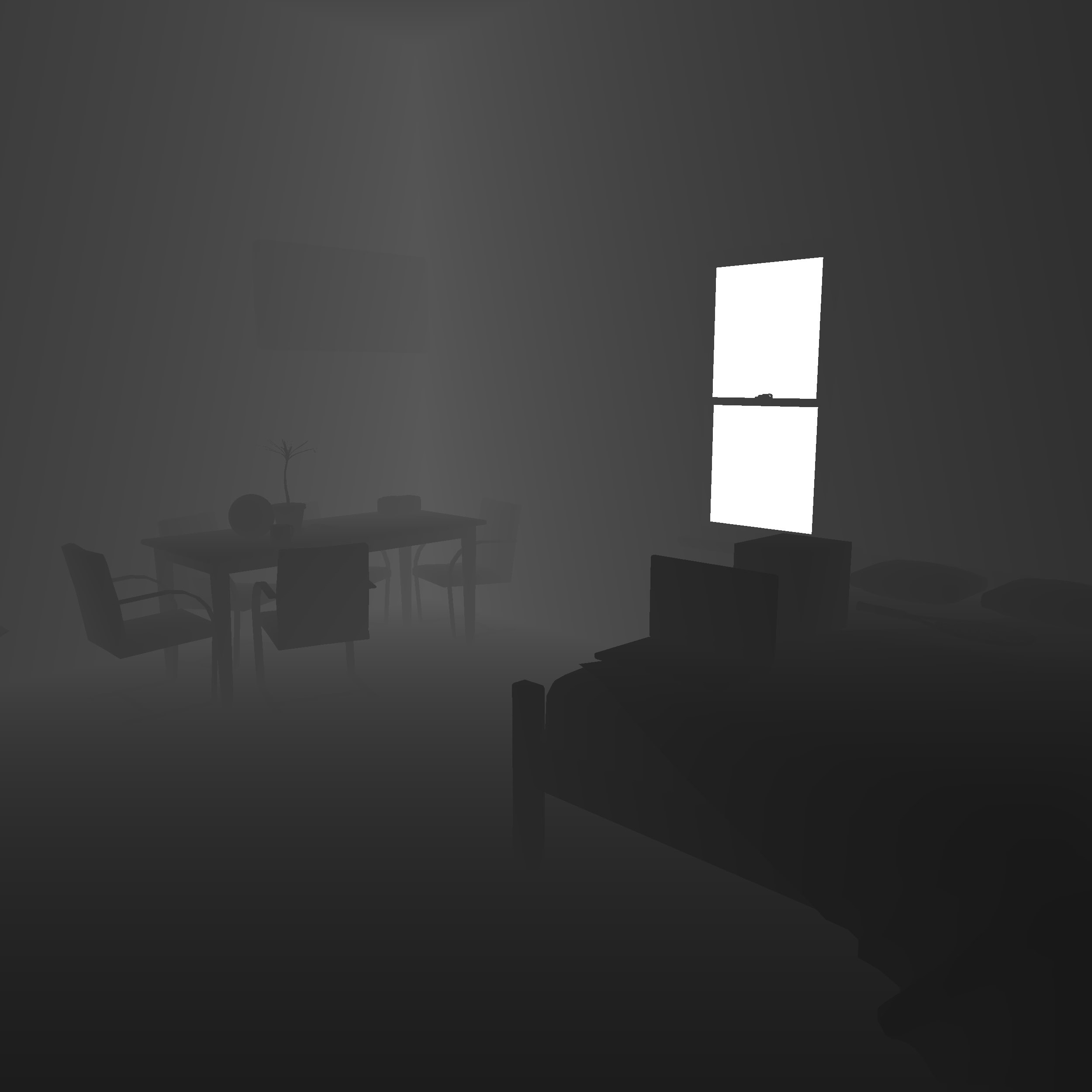}
         \caption*{Depth}
     \end{subfigure}
     \hfill
     \begin{subfigure}[b]{0.195\textwidth}
         \centering
         \includegraphics[width=\textwidth]{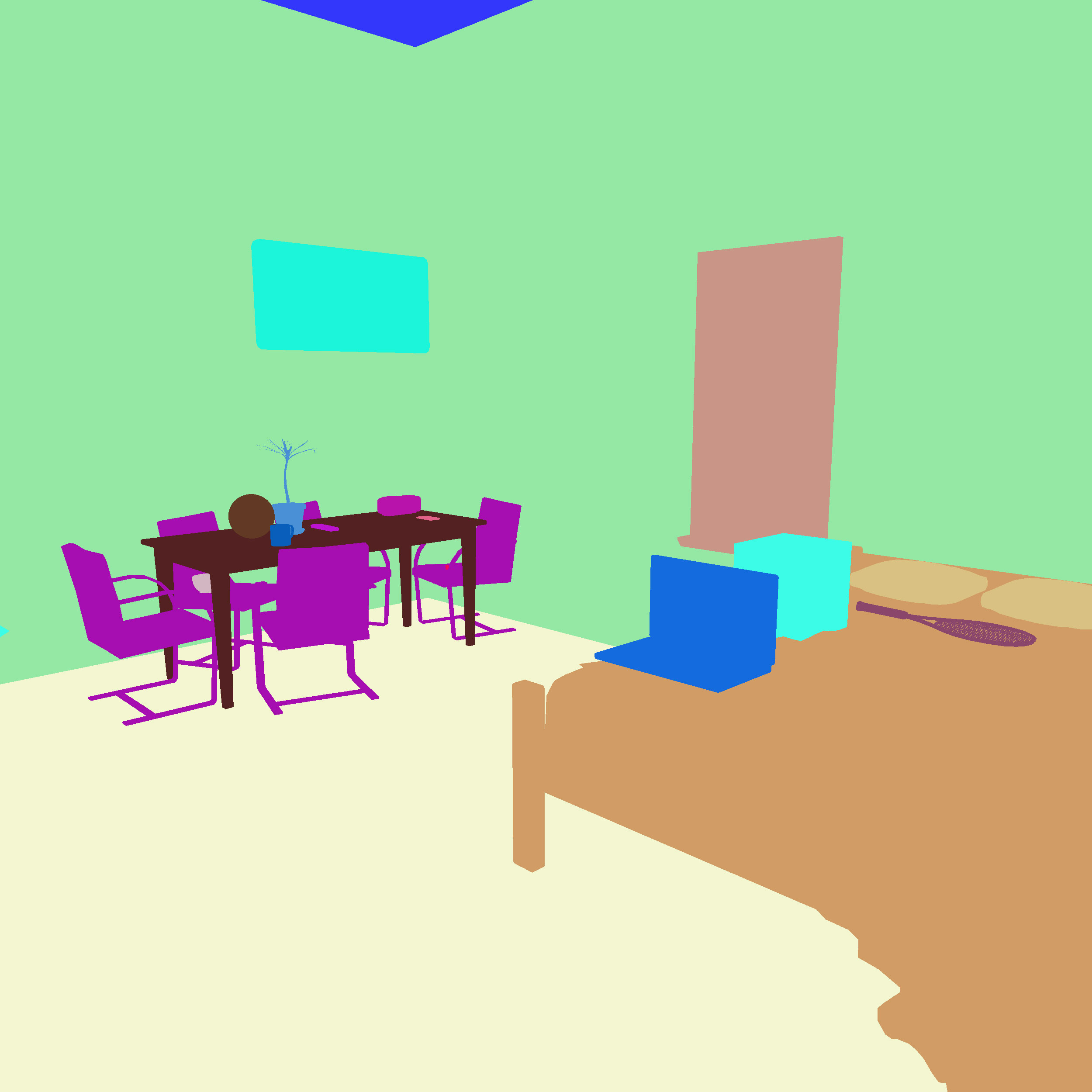}
         \caption*{Semantic Segmentation}
     \end{subfigure}
     \hfill
     \begin{subfigure}[b]{0.195\textwidth}
         \centering
         \includegraphics[width=\textwidth]{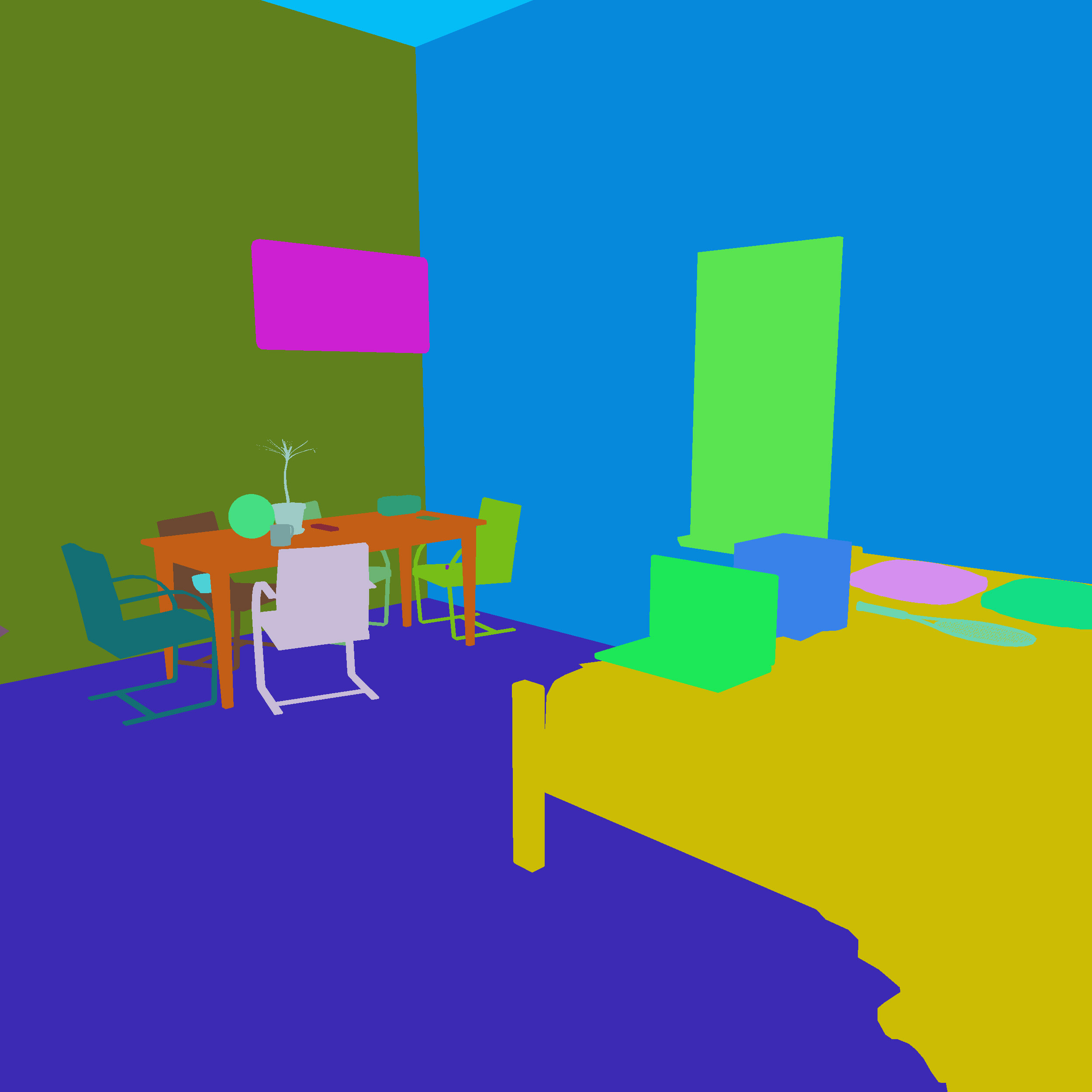}
         \caption*{Instance Segmentation}
     \end{subfigure}
     \hfill
     \begin{subfigure}[b]{0.195\textwidth}
         \centering
         \includegraphics[width=\textwidth]{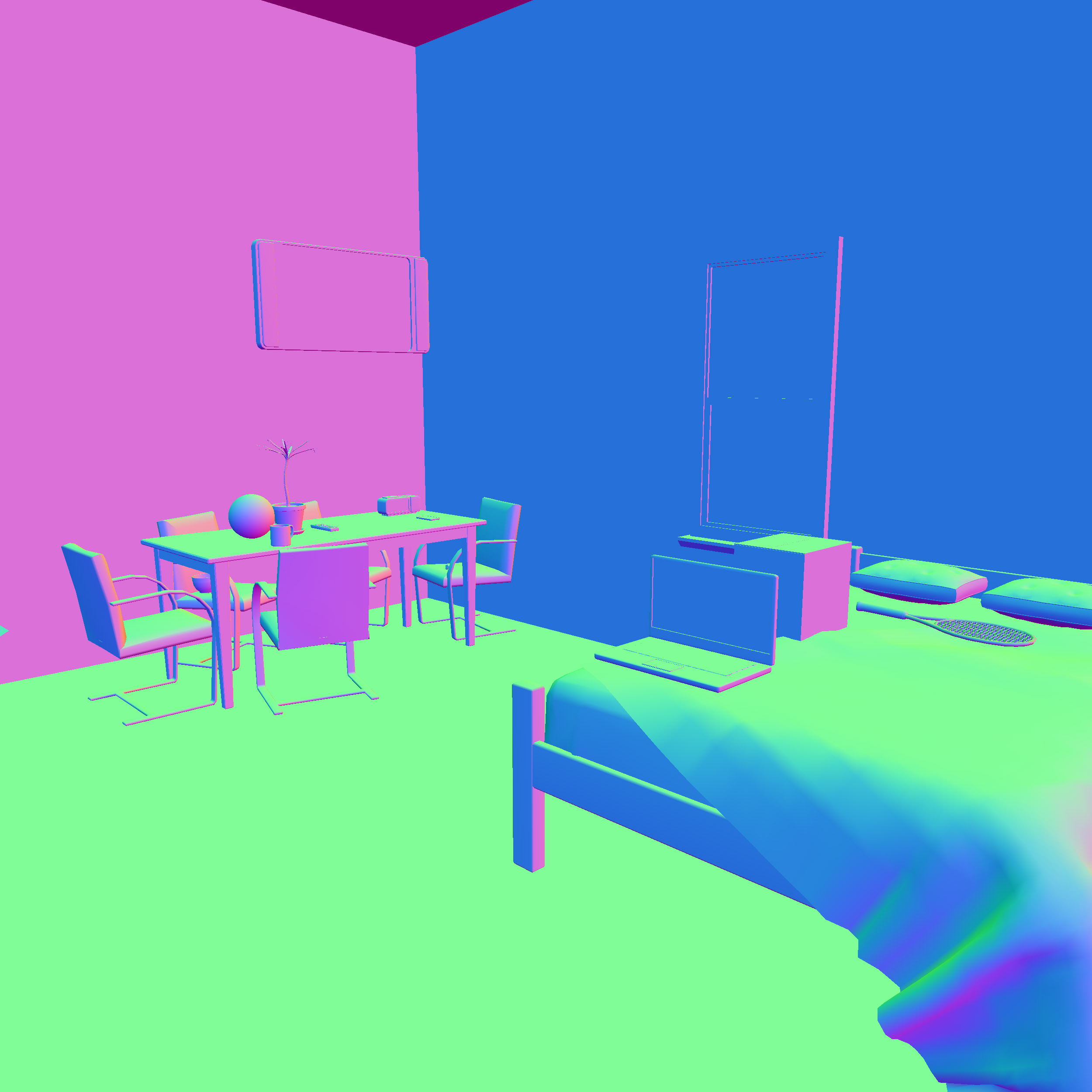}
         \caption*{Normals}
     \end{subfigure}
    \caption{Examples of image modalities supported in AI2-THOR, including RGB, depth, semantic segmentation, instance segmentation, and normals.}
    \label{fig:modalities}
\end{figure}

Figure~\ref{fig:modalities} shows a suite of different image modalities that can be rendered from each of the cameras in the scene, including RGB, depth, semantic segmentation, instance segmentation, and normals. Each agent comes with a camera attached to it, but more cameras can also be added, such as one to capture a top-down view of the scene. More image modalities can be added by modifying the Unity back-end (often by adding shaders).

\subsection{Objects}

\begin{figure}[ht!]
    \centering
    \includegraphics[width=\textwidth]{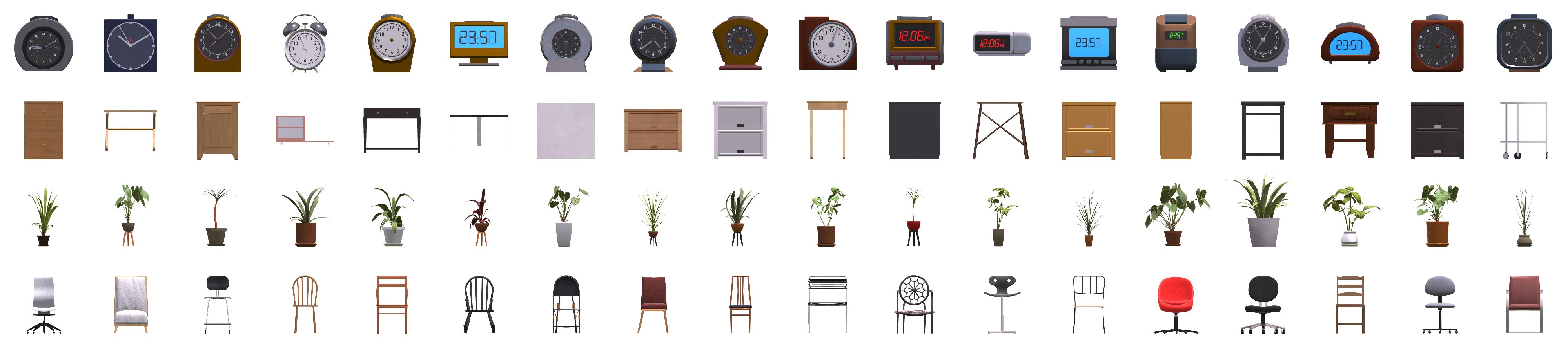}
    \begin{center}
        $\cdots$
    \end{center}
    \caption{Examples of objects in AI2-THOR's object database.}
    \label{fig:objects}
\end{figure}

AI2-THOR includes 3,578 interactive objects in its object database, which is rapidly growing. Each of these objects has been hand-modeled to support our set of interactive actions and state changes, such as opening, breaking, or cooking. Figure \ref{fig:objects} shows samples of objects from 4 categories, including alarm clocks, side tables, plants, and chairs.

\subsection{Environment Metadata}

\begin{figure}[ht!]
    \centering
    \begin{subfigure}[b]{0.32\textwidth}
        \centering
        \includegraphics[width=\textwidth]{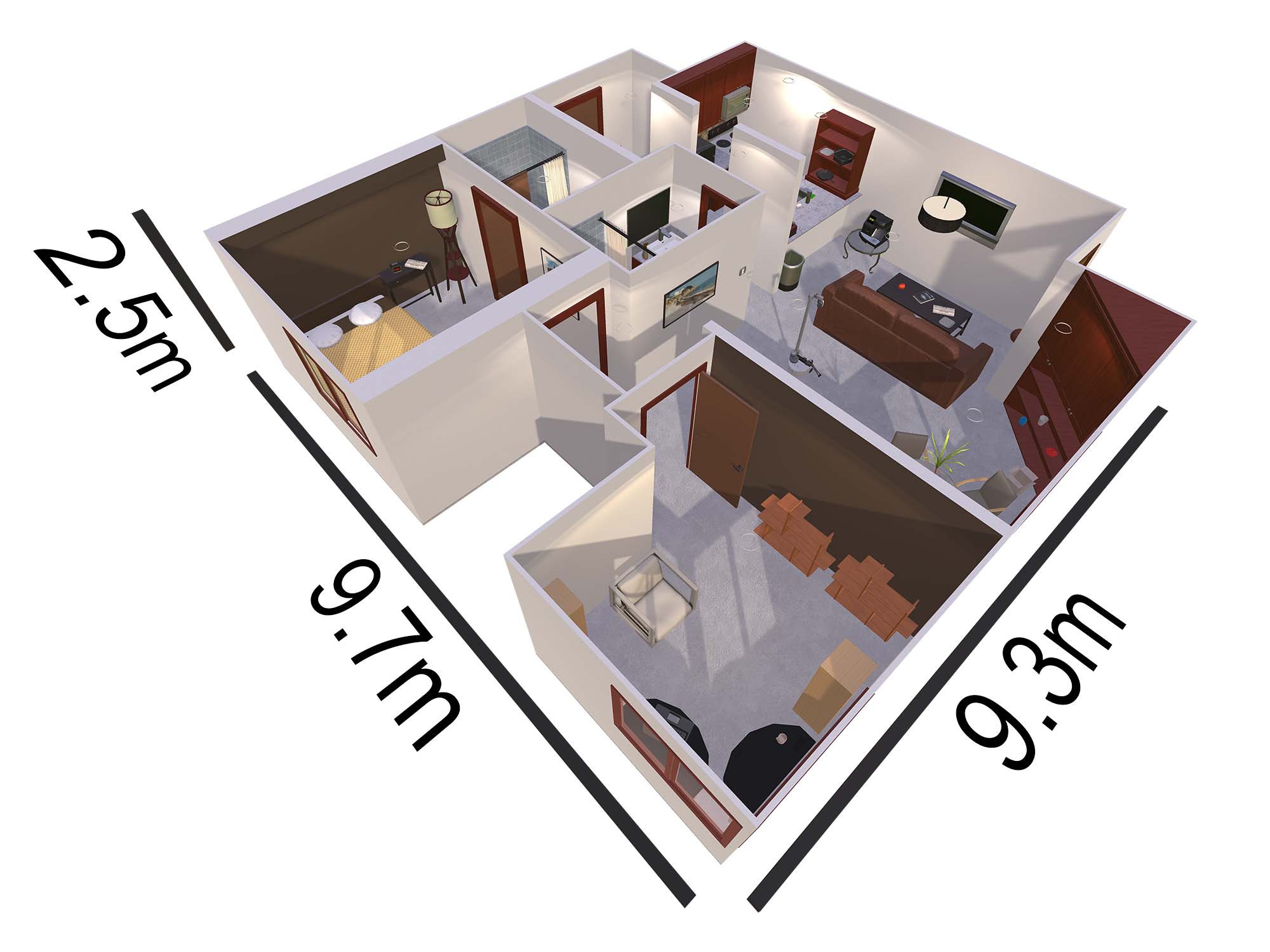}
        \caption{Scene Bounds}
    \end{subfigure}
    \hfill
    \begin{subfigure}[b]{0.32\textwidth}
        \centering
        \includegraphics[width=\textwidth]{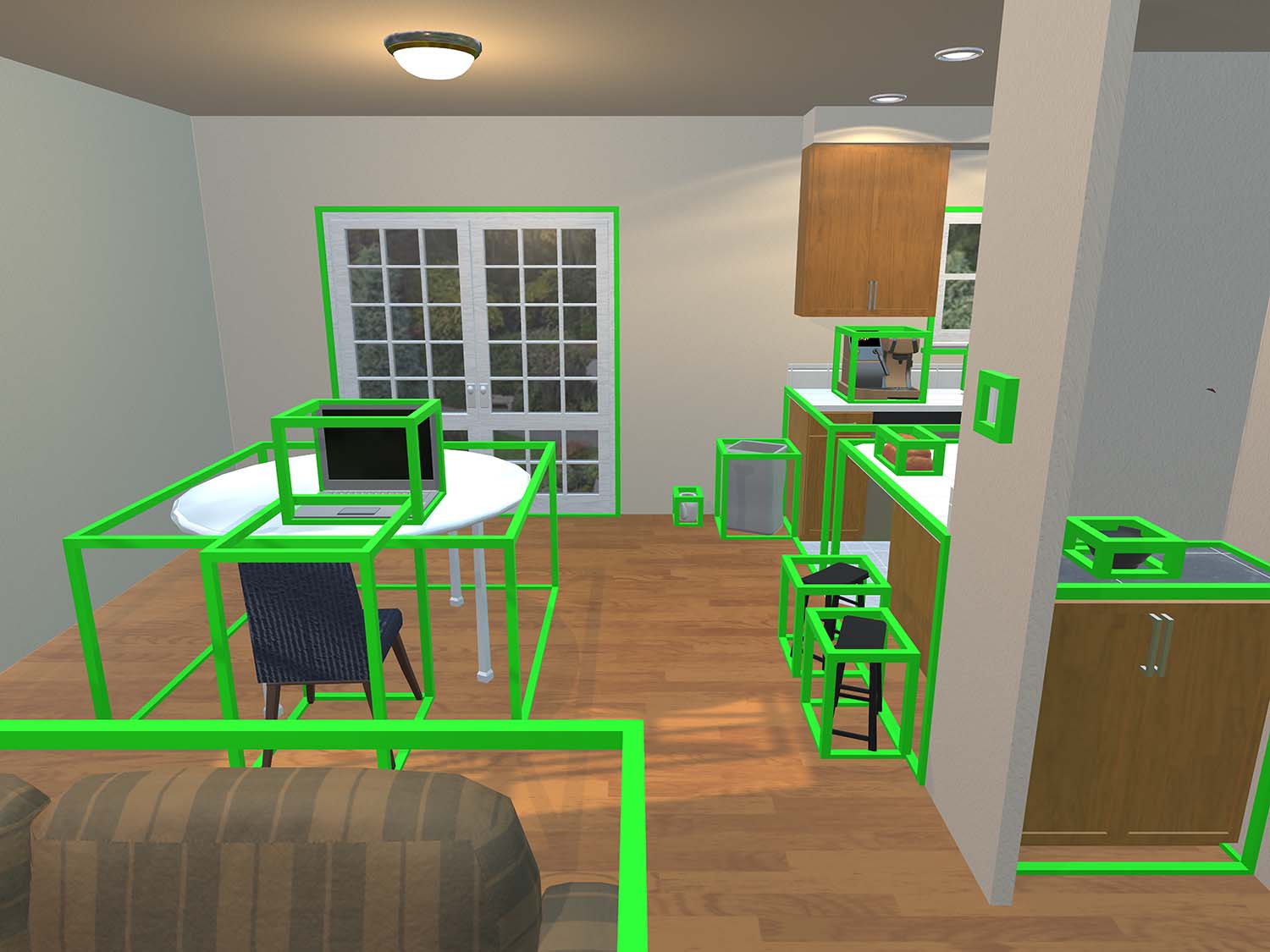}
        \caption{3D Bounding Box of Objects}
    \end{subfigure}
    \hfill
    \begin{subfigure}[b]{0.32\textwidth}
       \centering
        \includegraphics[width=\textwidth]{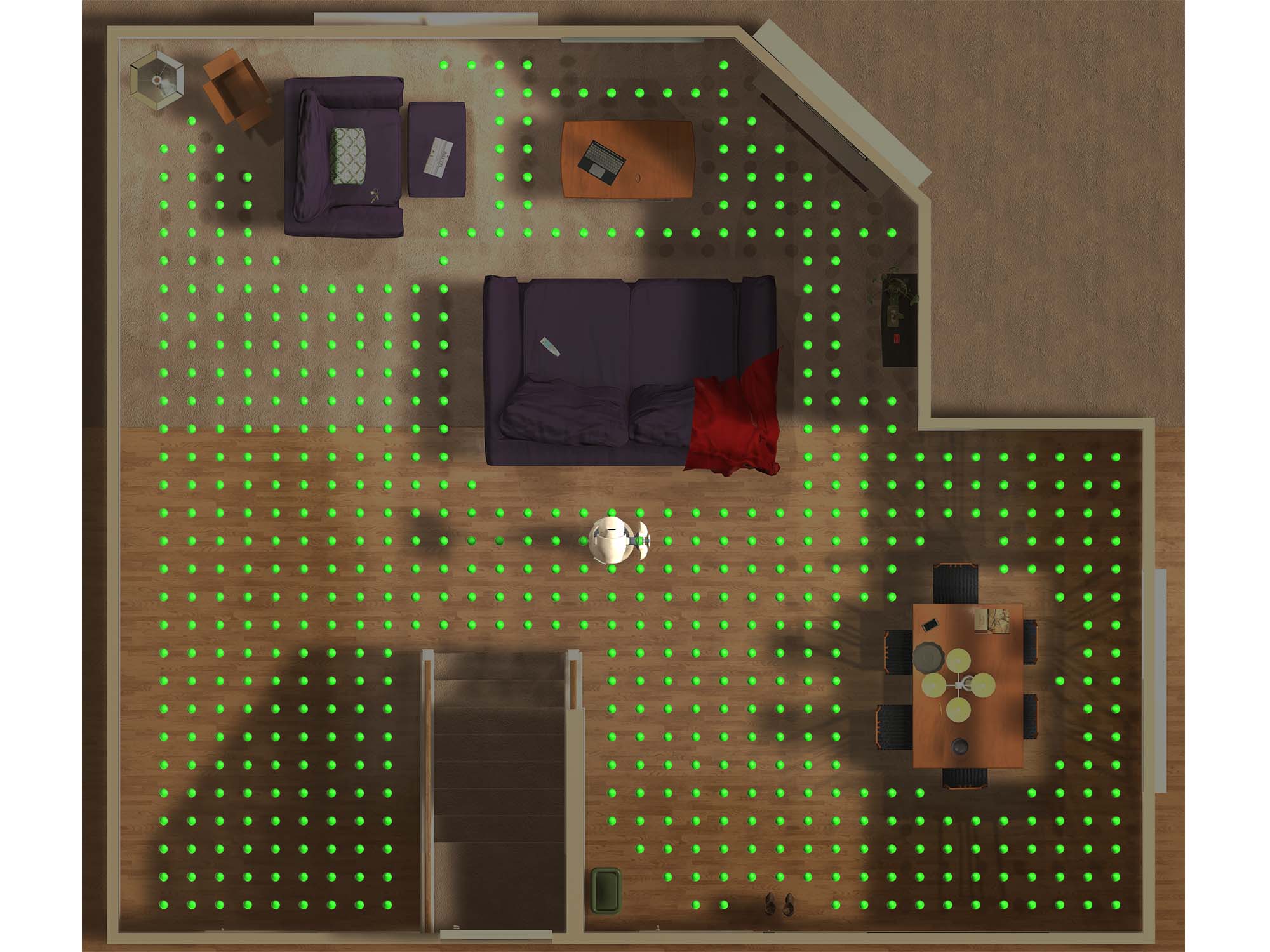}
        \caption{Reachable Grid Positions}
    \end{subfigure}
    \caption{Examples of environment metadata, including the dimensions of the scene, the 3D bounding box of each object, and the reachable grid positions, which may be used to randomize the agent's starting position or to build a heuristic search agent.}
    \vspace{0.3cm}
    \label{fig:environmentMetadata}
\end{figure}

Environment metadata is returned after each action is executed. It includes information such as the pose of each agent; the pose and state of each object in the scene (\eg, whether the object is moving, if it is visible to the agent, how far open it is, if it is clean or dirty); metadata about the scene, such as its size; and if the most recent action executed successfully (\eg, the agent did not collide with an object while trying to move). Metadata is often not provided to the agent for most tasks, as it would make the tasks too simple and easily solvable with a heuristic. Instead, many tasks use metadata to build a reward function with access to ``expert-level'' information that is hidden from the agent, build an imitation learning expert, and construct training and evaluation datasets.

\section{What has AI2-THOR been used for?}

\begin{figure}[ht!]
    \makebox[\linewidth][c]{
        \begin{subfigure}[b]{0.3\textwidth}
            \centering
            {\small\textbf{Task:} Find the Bed}\\[-0.1in]
            \includegraphics[width=\textwidth]{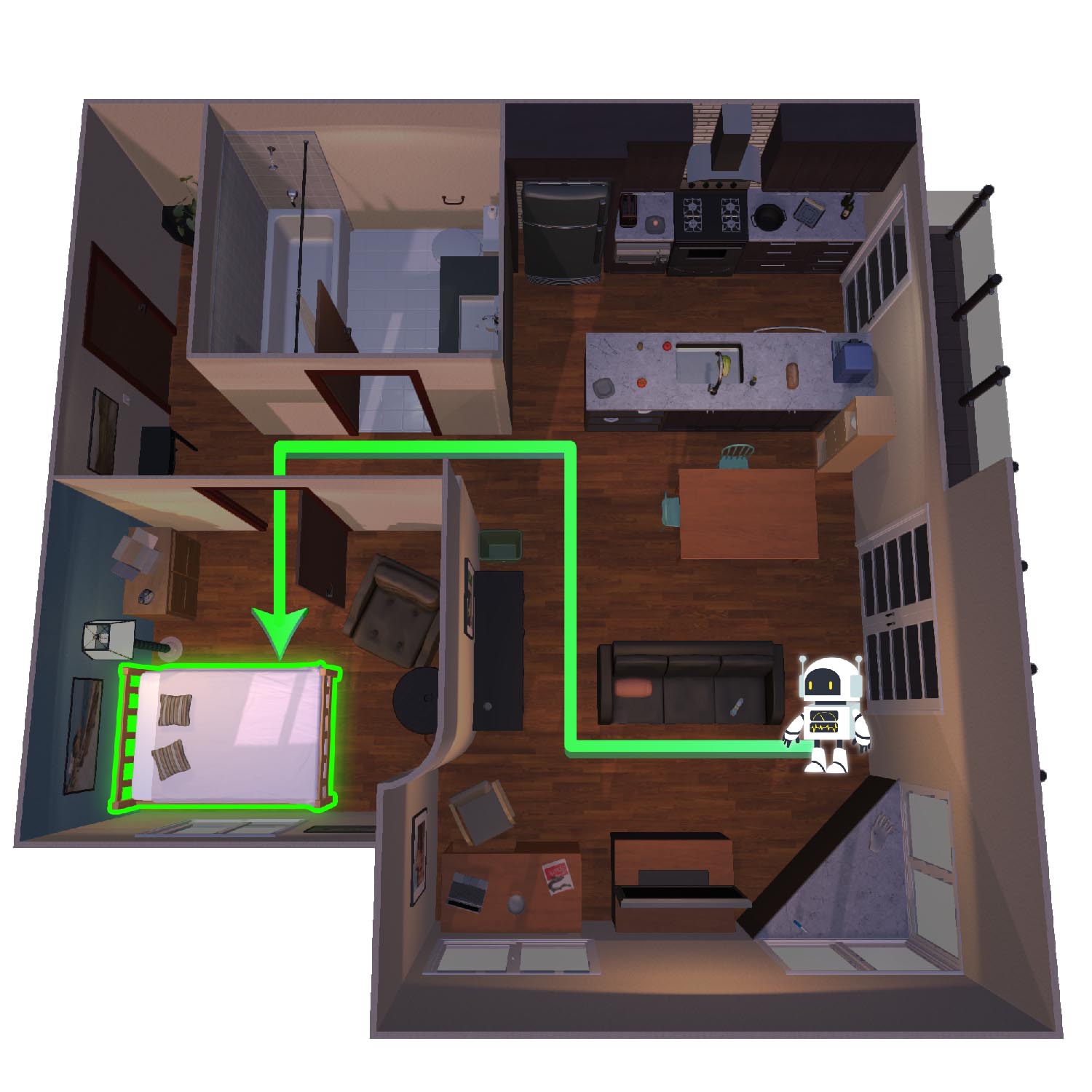}
            \vspace{-0.25in}
            \caption*{Visual Navigation~\cite{zhu2017target}}
        \end{subfigure}
        \hfill
        \begin{subfigure}[b]{0.3\textwidth}
            \centering
            {\small\textbf{Task:} Find the Sound}\\[0.025in]
            \includegraphics[width=0.98\textwidth]{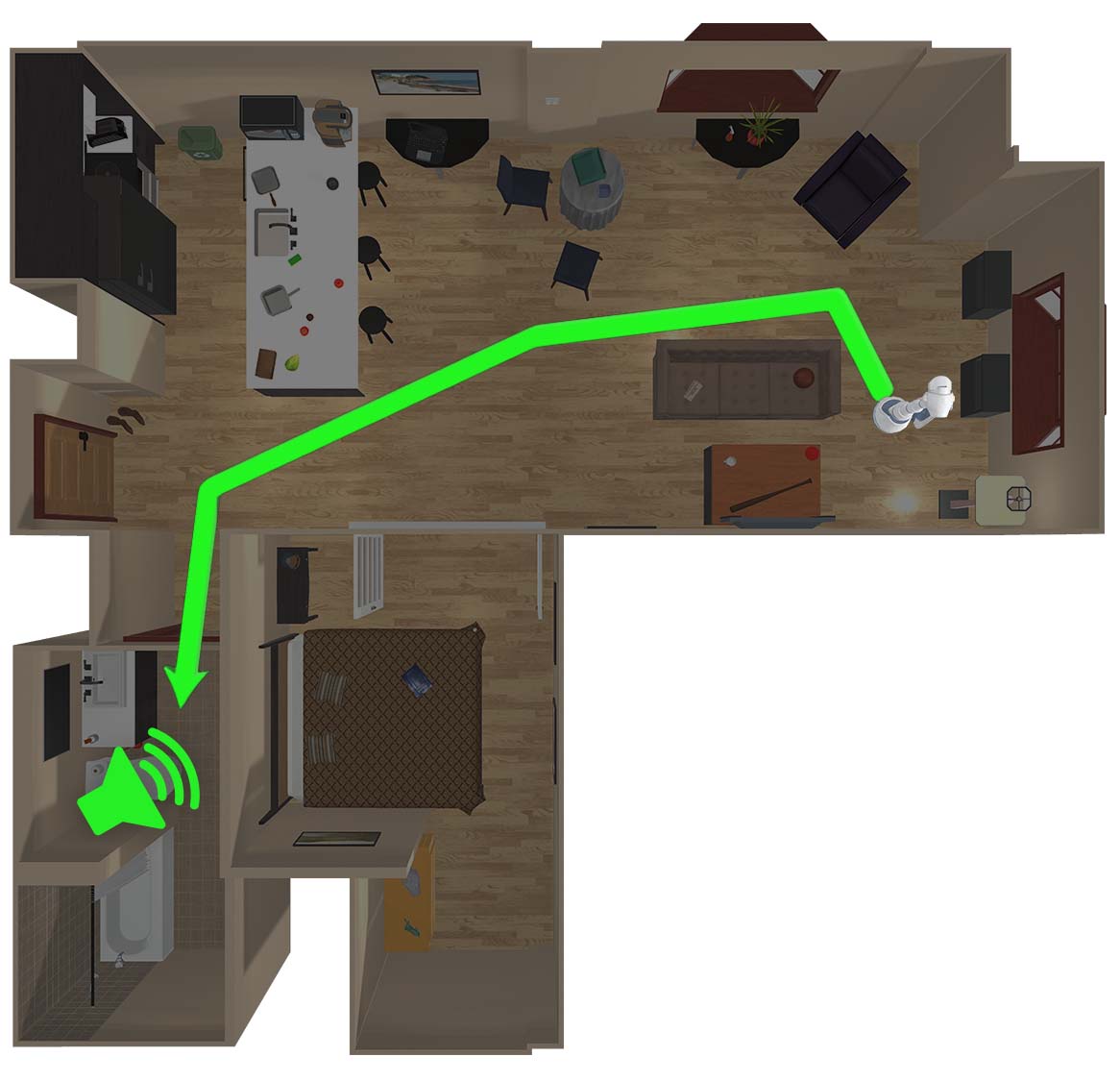}
            \caption*{Audio-Visual Navigation~\cite{gan2020look}}
        \end{subfigure}
        \hfill
        \begin{subfigure}[b]{0.3\textwidth}
            \centering
            \begin{overpic}[width=\textwidth]{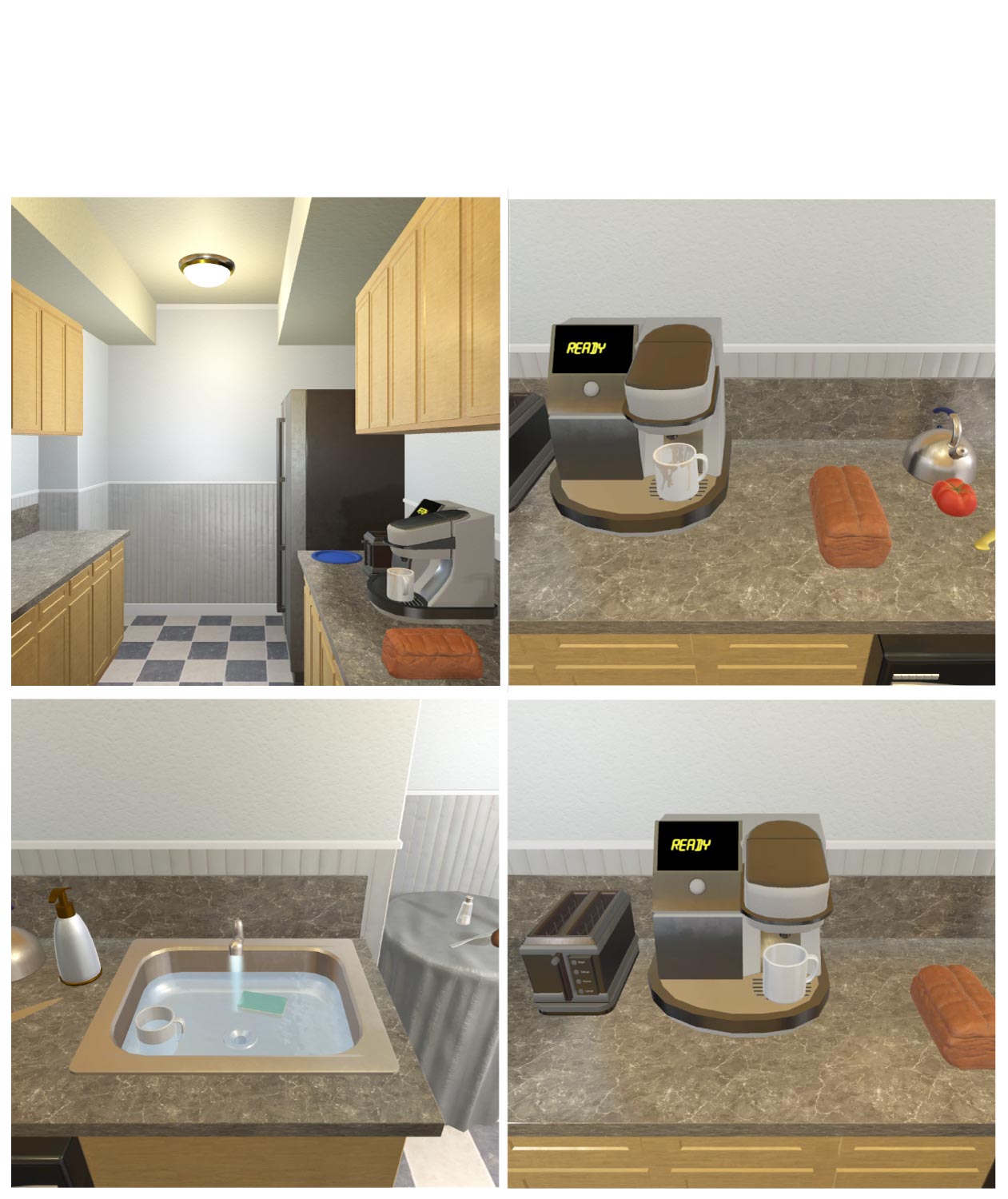}
                \put(0,0){\includegraphics[width=1\textwidth]{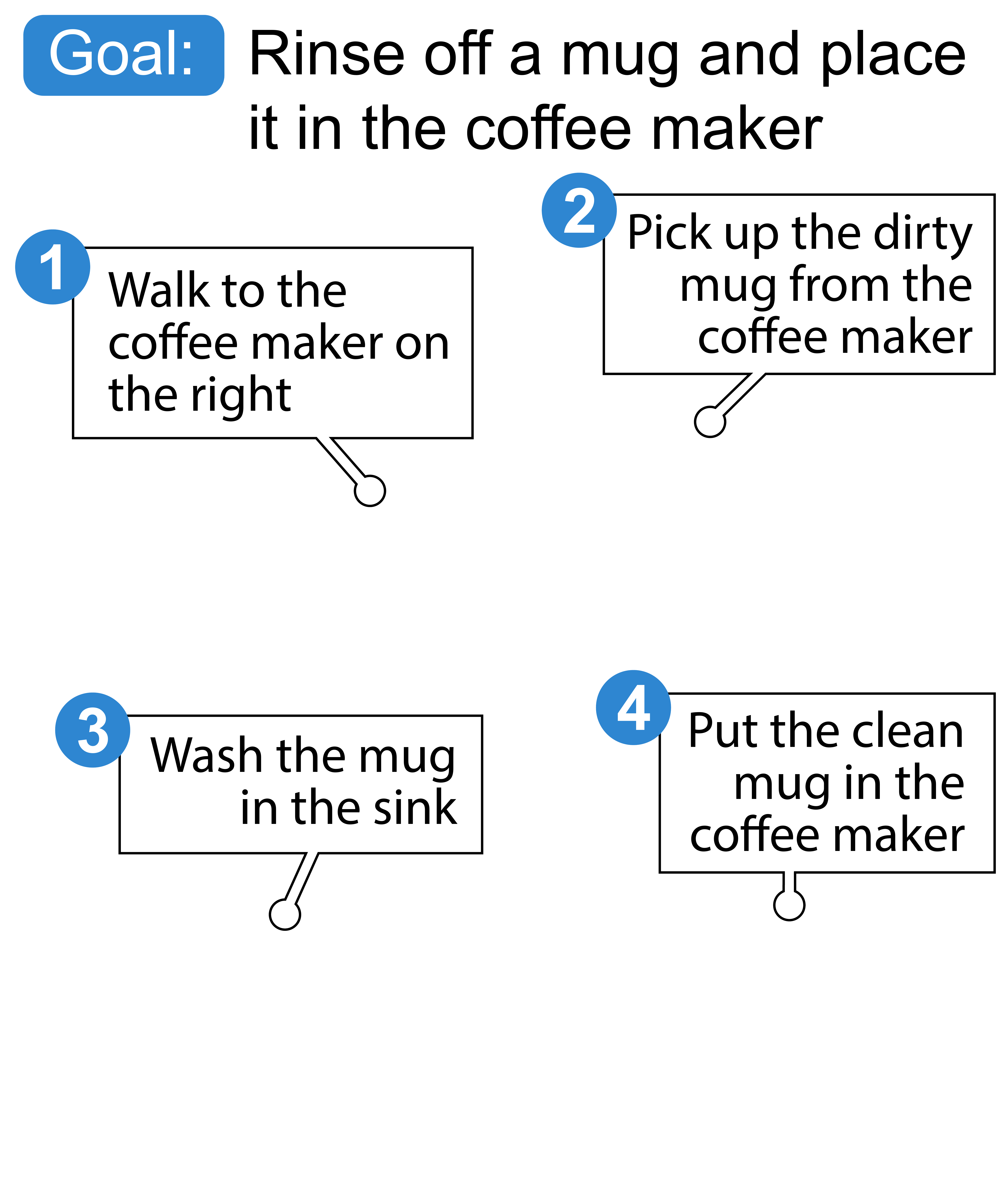}}
            \end{overpic}
            \caption*{Vision-and-Language~\cite{shridhar2020alfred}}
        \end{subfigure}
        \hfill
        \begin{subfigure}[b]{0.3\textwidth}
            \centering
            \includegraphics[width=\textwidth]{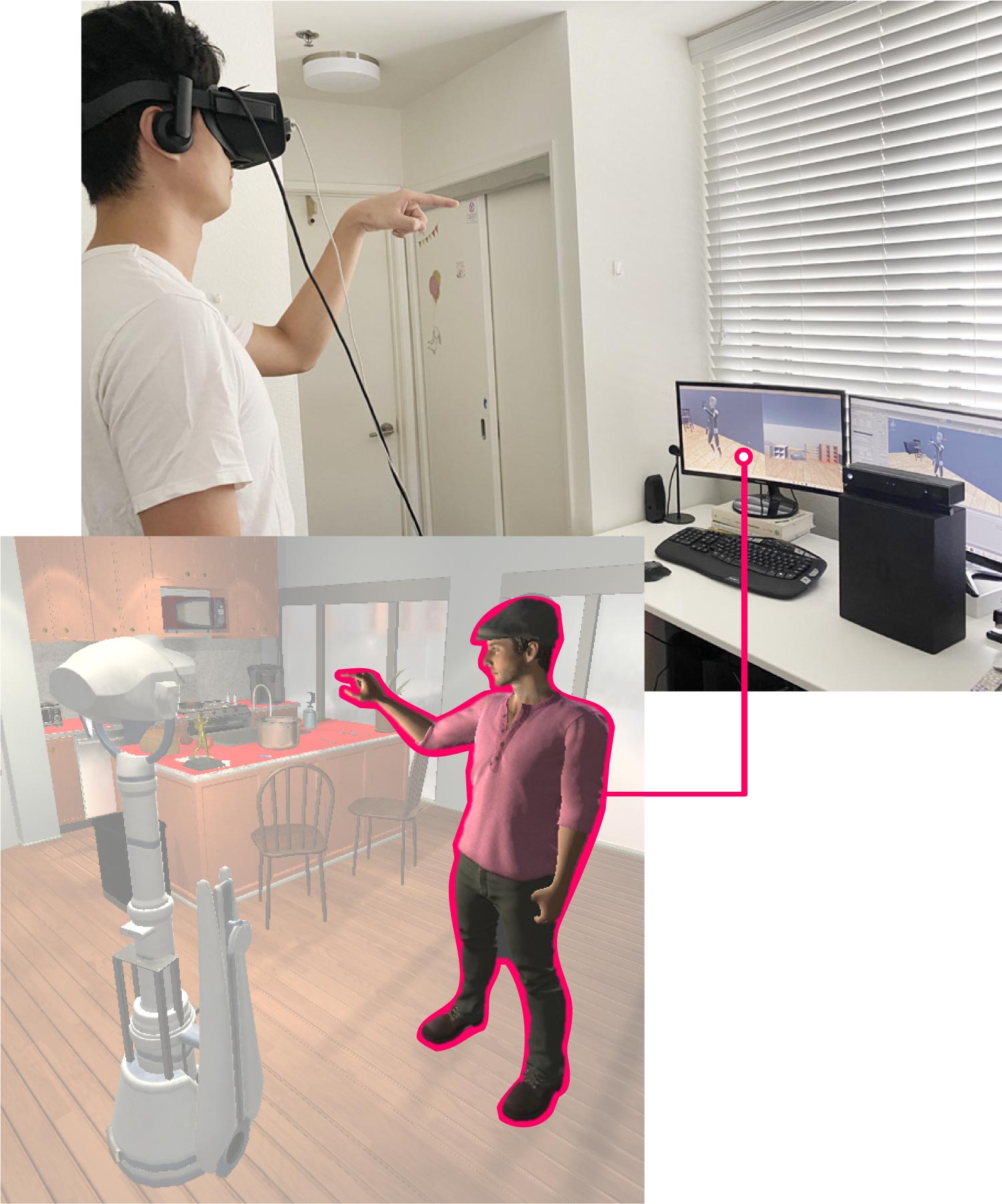}
            \caption*{Human Robot Interaction~\cite{wu2021communicative}}
        \end{subfigure}
    }\\[0.2in]
    \makebox[\linewidth][c]{
        \begin{subfigure}[b]{0.3\textwidth}
            \centering
            \includegraphics[width=\textwidth]{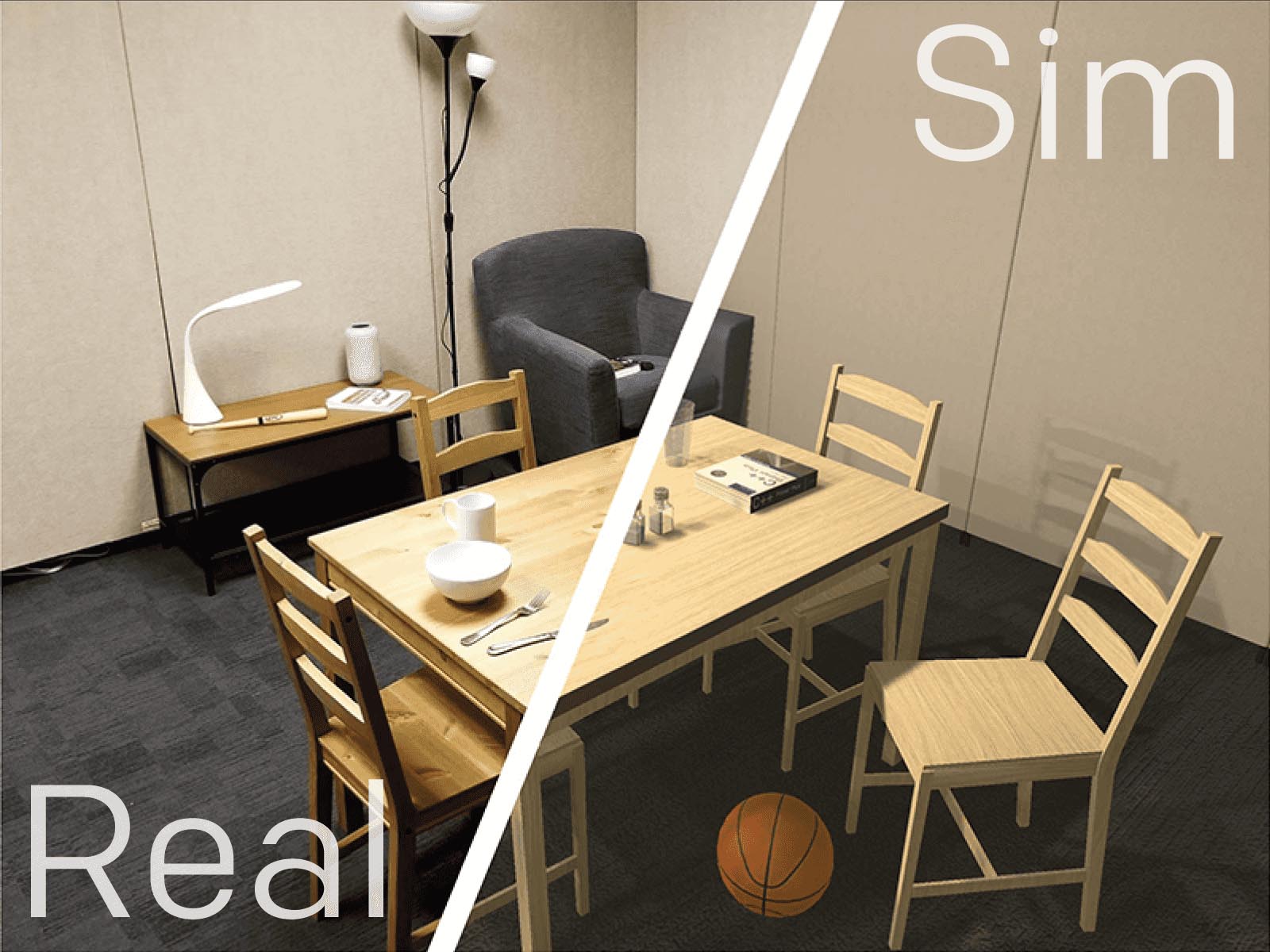}
            \caption*{Sim2Real Robotics~\cite{deitke2020robothor}}
        \end{subfigure}
        \hfill
        \begin{subfigure}[b]{0.3\textwidth}
            \centering
            \includegraphics[width=\textwidth]{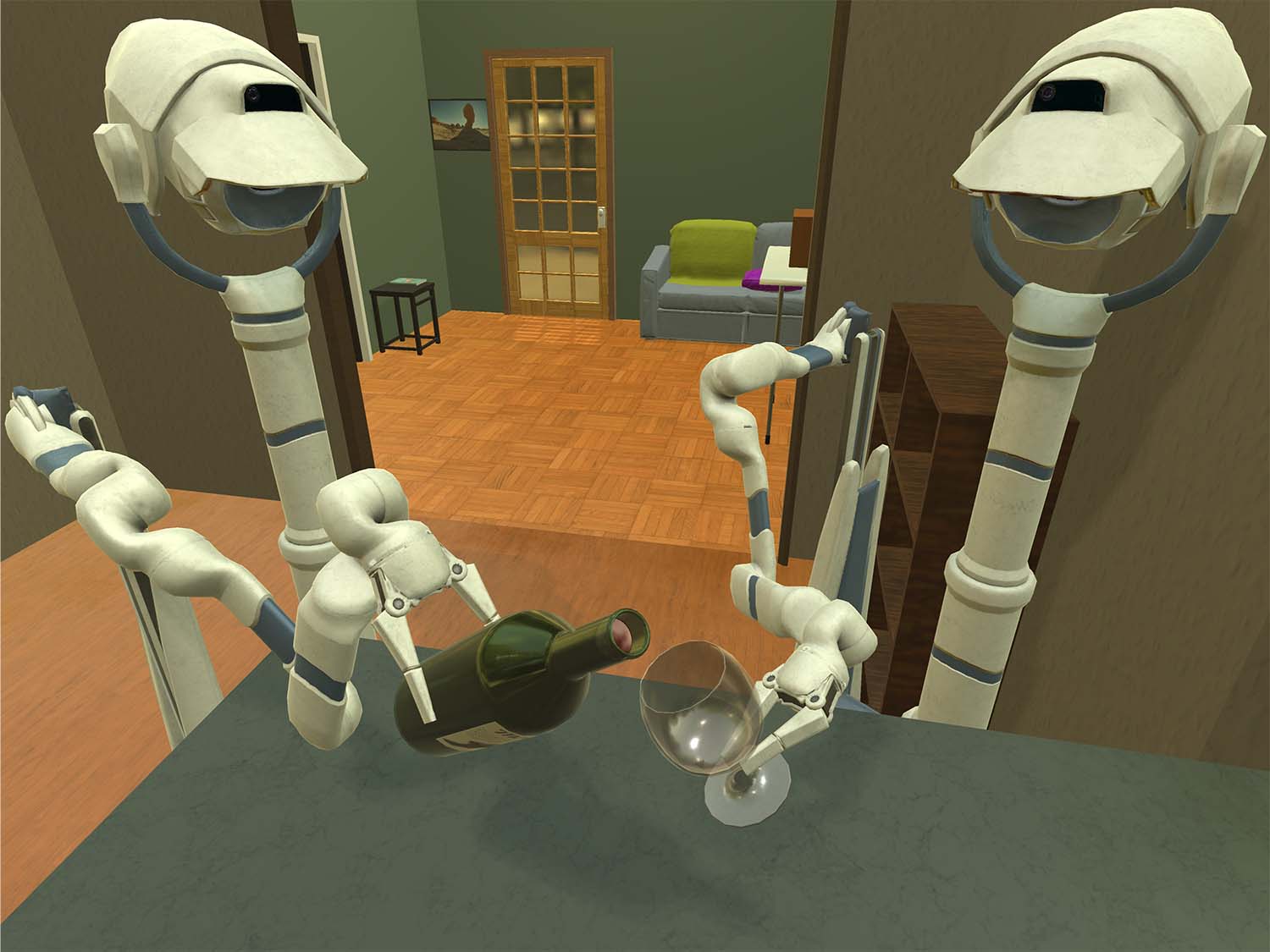}
            \caption*{Multi-Agent Interaction~\cite{TwoBody}}
        \end{subfigure}
        \hfill
        \begin{subfigure}[b]{0.3\textwidth}
            \centering
            \includegraphics[width=\textwidth]{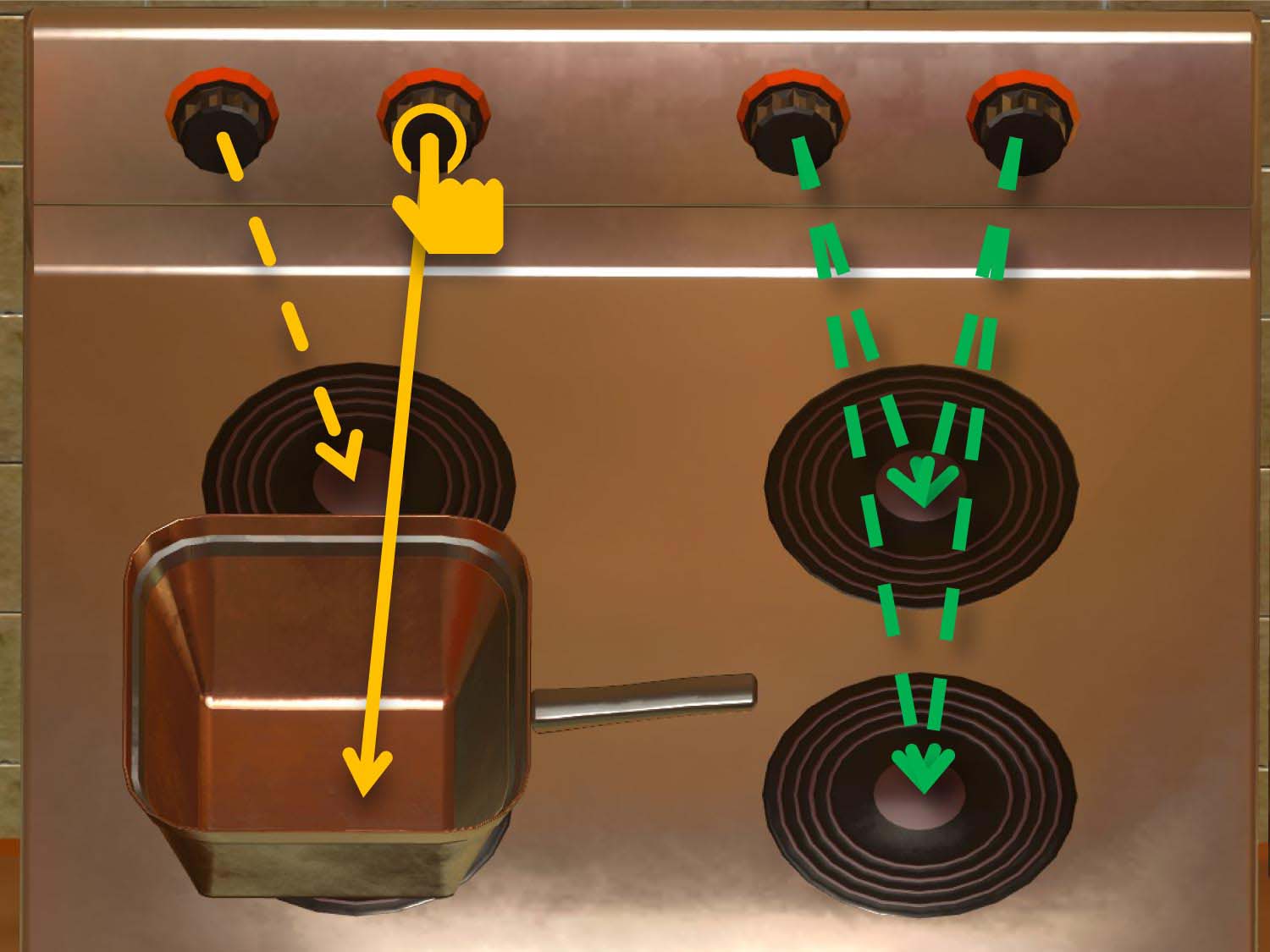}
            \caption*{Learning Object Relationships~\cite{li2021ifr}}
        \end{subfigure}
        \hfill
        \begin{subfigure}[b]{0.3\textwidth}
            \centering
            \begin{overpic}[width=\textwidth]{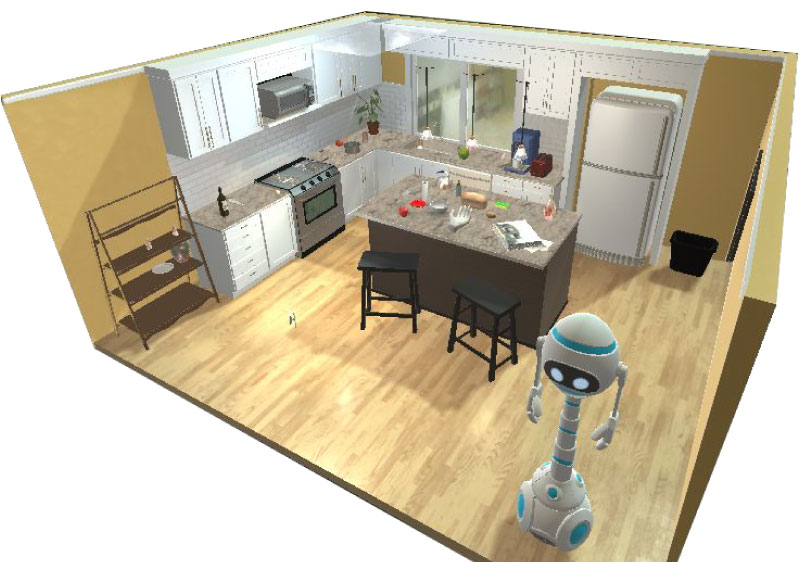}
                \put(0,0){\includegraphics[width=1\textwidth]{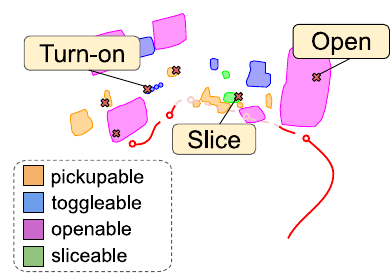}}
            \end{overpic}
            \caption*{Learning Affordances~\cite{nagarajan2020learning}}
        \end{subfigure}
    }\\[0.15in]
    \makebox[\linewidth][c]{
        \begin{subfigure}[b]{0.3\textwidth}
            \centering
            \includegraphics[width=\textwidth]{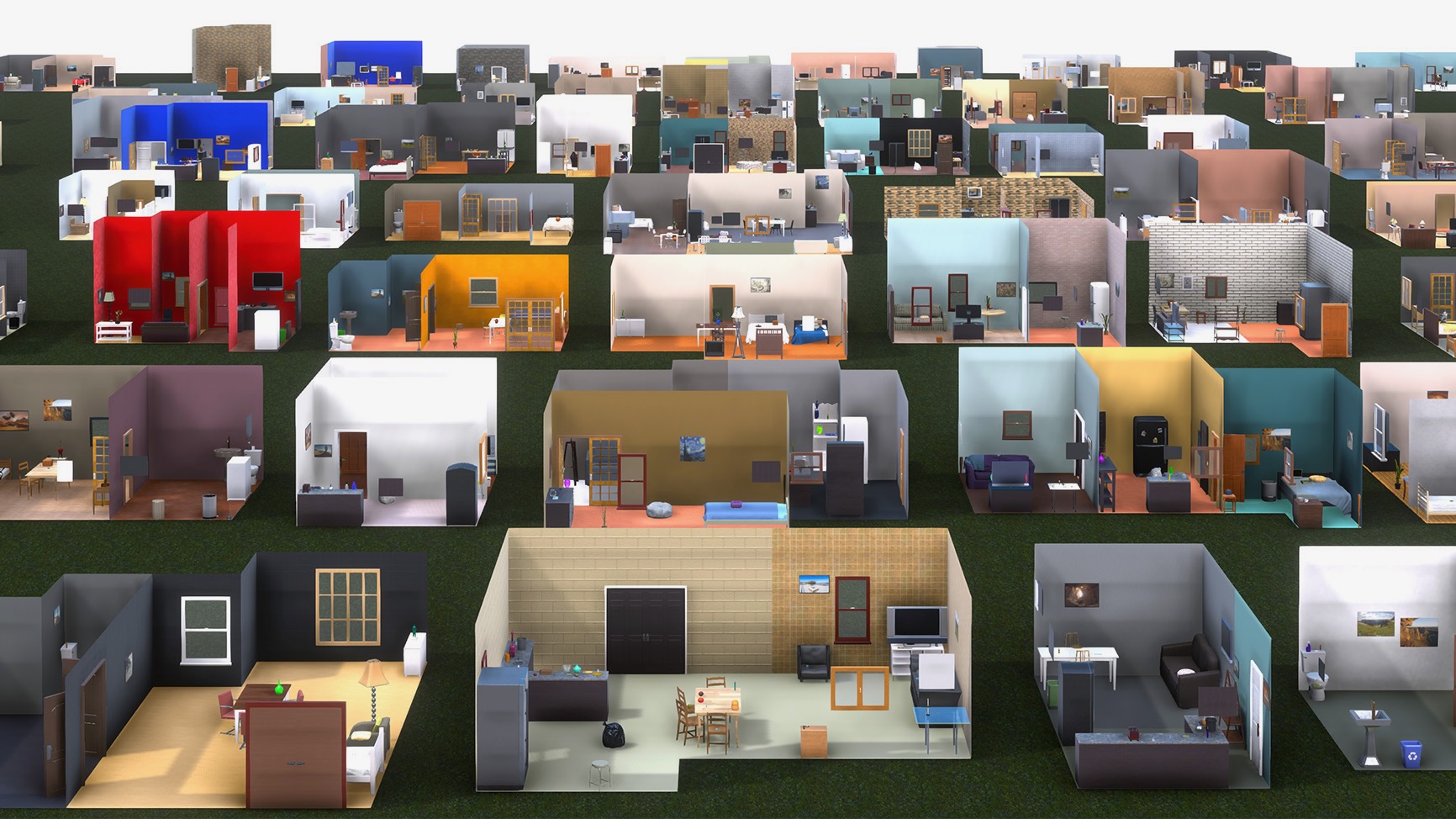}
            \caption*{Scene Synthesis~\cite{deitke2022procthor}}
        \end{subfigure}
        \hfill
        \begin{subfigure}[b]{0.3\textwidth}
            \centering
            \includegraphics[width=\textwidth]{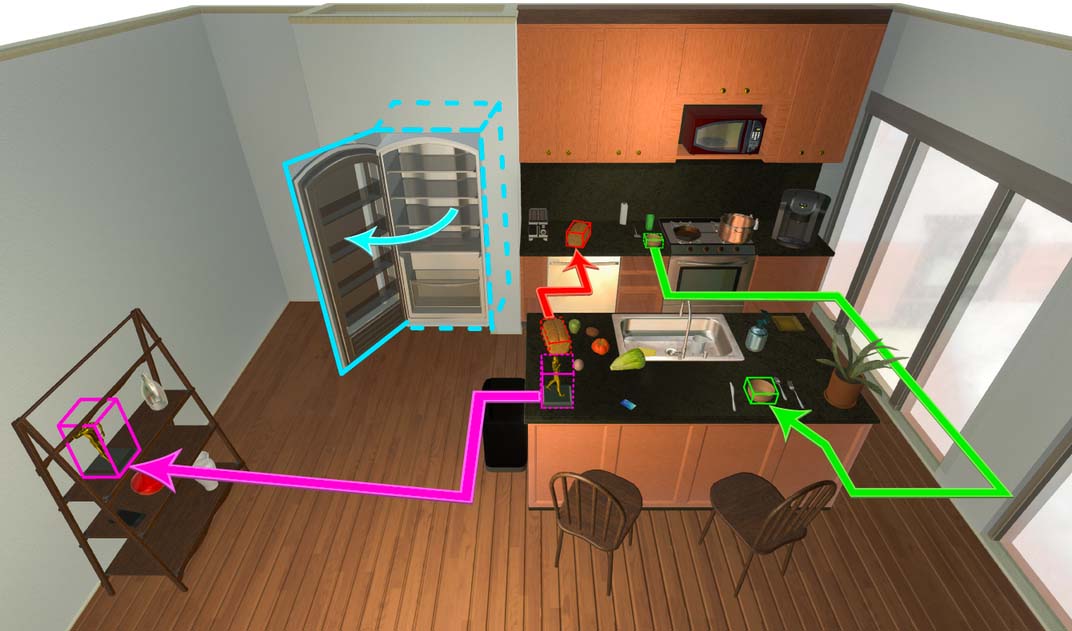}
            \caption*{Learning with Interaction~\cite{weihs2021visual}}
        \end{subfigure}
        \hfill
        \begin{subfigure}[b]{0.3\textwidth}
            \centering
            \includegraphics[width=\textwidth]{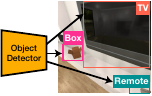}
            \caption*{Computer Vision~\cite{kotar2022interactron}}
        \end{subfigure}
        \hfill
        \begin{subfigure}[b]{0.3\textwidth}
            \centering
            \includegraphics[width=\textwidth]{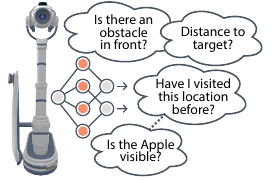}
            \caption*{Interpretability~\cite{dwivedi2022navigation}}
        \end{subfigure}
    }\\[-0.05in]
    \caption{AI2-THOR has enabled research in a wide range of fields. Here, we highlight some examples of how it has been used.}
\end{figure}

Since the initial release of AI2-THOR in 2017, it has been used for experimentation in over 150 publications and downloaded over 500k times. Some areas of work that we found particularly interesting include:
\begin{itemize}[leftmargin=0.25in]
    \item \textbf{Visual Navigation.} Visual navigation was the first use case of AI2-THOR~\cite{zhu2017target}, which trains an agent to perform ImageNav (\ie navigating to an image where the target object is described with a picture of it). Here, the agent executes a sequence of move or rotate commands to reach the target from egocentric camera inputs at each time step. ObjectNav is another common navigation task, where the agent is tasked with navigating to a given semantic category, such as a bed. Follow-up work from \cite{yang2018visual, du2020learning,zheng2022towards} uses semantic priors about where objects typically occur to improve navigation efficiency; \cite{wortsman2019learning} used meta-learning to try and better adapt to unseen scenes; \cite{lu2021mgrl} uses a Markov network to build a map of the environment; \cite{khandelwal2022simple} found that using CLIP as a pre-trained visual encoder helps significantly boost generalization performance; and \cite{deitke2022procthor} found that training on many procedurally generated scenes strongly generalizes to RoboTHOR, iTHOR, and ArchitecTHOR in a 0-shot setting.
    \item \textbf{Audio-Visual Navigation.} \cite{gan2020look} proposes the task of audio-visual navigation in which the agent is tasked with navigating to find where the sound is coming from in the scene.
    \item \textbf{Vision-and-Language.} AI2-THOR has been used extensively for embodied vision-and-language research. Noteable datasets include ALFRED~\cite{shridhar2020alfred}, for interactive instruction following from natural language; TEACh~\cite{padmakumar2022teach}, for interactive instruction following from human-robot dialog; and DialFRED~\cite{gao2022dialfred} and IQA~\cite{gordon2018iqa} for interactive question-answering. Some other interesting work includes \cite{pashevich2021episodic}, which proposes the Episodic Transformer to encode the full history of vision and language inputs with each ALFRED task; \cite{karamcheti2020learning}, which uses grammar-based methods to learn high-level abstractions through decompositions of tasks; FILM~\cite{min2021film}, which builds a semantic map to perform exploration for instruction following; and PIGLeT~\cite{zellers2021piglet}, which learns natural language grounding through interaction.
    \item \textbf{Human-Robot Interaction.} \cite{wu2021communicative} inserts a human into AI2-THOR and uses virtual reality to control its gestures in simulation. By controlling the human's gestures, it can communicate different tasks it wants the robot to achieve, such as pointing to an object to encode moving to that object.
    \item \textbf{Sim2Real Transfer.} RoboTHOR~\cite{deitke2020robothor} studies sim2real transfer for robotics. Here, the goal is to train in simulation because it is faster, cheaper, and more scalable, and then to deploy the trained agent in the real-world. Agents train on 75 scenes in simulation and evaluate on unseen real-world scenes that come from a similar distribution. Initial work analyzed sim2real transfer for agents trained to perform ObjectNav.
    \item \textbf{Multi-Agent Interaction.} \cite{TwoBody} proposes the collaborative task of having 2 agents move to lift up furniture in a scene. For example, both agents might have to navigate to find the television in the scene, and work together to lift it up. Follow-up work from \cite{CordialSync} takes the task a step further, where the agents not only have to lift up the furniture, but also work together to move it. Both tasks require visual navigation from the agents, and for them to communicate and coordinate together. Some other notable multi-agent work includes \cite{weihs2019learning}, which tasks agents with playing Cache, a variant of hide-and-seek where one agent hides an object and the other agent is tasked with finding that object; \cite{tan2020multi}, which uses multiple agents for interactive question answering; \cite{liu2022multi}, which proposes using multiple agents to more efficiently find multiple target objects in a scene; and TEACh~\cite{padmakumar2022teach}, which uses a commander agent and a follower agent to mimic human-robot dialog to solve interactive tasks.
    \item \textbf{Learning Object Relationships.} \cite{li2021ifr} proposes an approach to learn priors about inter-object functional relationships, such as which knobs on the stove control each burner, that the light switches controls a given light, and that the remote may control a television. \cite{nagarajan2021shaping} proposes using egocentric videos to learn which objects are used together to complete certain activities. They then use the priors to help guide agents towards achieving different activities in AI2-THOR.
    \item \textbf{Learning Affordances.} \cite{nagarajan2020learning} train an agent to interact with the environment to learn object affordances, which encode which objects may be interacted with and how. For instance, it learns that drawers or fridges may be opened, that the stove can turn on, and that an apple may be sliced. A model with an affordance landscape would make it easier to adapt to downstream tasks, such as learning to cut a tomato with a knife.
    \item \textbf{Scene Synthesis.} ProcTHOR~\cite{deitke2022procthor} uses procedural generation to synthesize training houses at scale to improve the generalization abilities of embodied agents. It procedurally generated and trained on 10K houses by first sampling floorplans and then plausibly placing objects within each of the rooms in the floorplan. Remarkably, pre-training on ProcTHOR alone was able to achieve state-of-the-art performance for ObjectNav on RoboTHOR, iTHOR, and ArchitecTHOR, without leveraging any additional training data. LUMINOUS~\cite{zhao2021luminous} also uses scene synthesis techniques to train embodied agents, where it focuses on placing objects in iTHOR rooms.
    \item \textbf{Learning with Interaction.} AI2-THOR supports a wide range of interactions that can be used to train agents, including for rearranging objects in a scene with RoomR~\cite{weihs2021visual}, arm-based manipulation with ManipulaTHOR~\cite{ehsani2021manipulathor}, learning about objects by interacting with them~\cite{lohmann2020learning}, and playing hide-and-seek with objects to learn visual representations~\cite{weihs2019learning}, among many others.
    \item \textbf{Computer Vision.} The rich annotations available in simulation make it easy to use AI2-THOR for pure computer vision tasks. Notable work includes SeGAN~\cite{ehsani2018segan}, which used a GAN to generate occluded parts of an object from images in scene; Interactron~\cite{kotar2022interactron}, which performs object detection with embodied agents that are able to move around in the environment; and \cite{kotar2021contrasting}, which performs depth estimation and action prediction to evaluate contrastive learning approaches.
    \item \textbf{Interpretability.} iSEE~\cite{dwivedi2022navigation} uses probing to discover what information is in the hidden representations of Embodied AI models. It focuses on probing ObjectNav and PointNav agents to answer interpretability questions, such as how far the agent thinks it is from the target.
\end{itemize}

AI2-THOR is rapidly updating to build out features and functionality. For the latest published papers, please visit the publication tracker on our website: \url{https://ai2thor.allenai.org/publications}.

\section{Why use AI2-THOR?}

\begin{table}[ht!]
    \centering
    \resizebox{1.0\textwidth}{!}{
    \begin{tabular}{l cc c ccccc c cc }
        \toprule
        & \multicolumn{2}{c}{Scale} && \multicolumn{5}{c}{Interaction} && \multicolumn{2}{c}{Simulator} \\
        \cmidrule{2-3}
        \cmidrule{5-9}
        \cmidrule{11-12}
        Simulator & \makecell{\# of\\Scenes} & \makecell{\# of\\Objects} && \makecell{Object\\States} & \makecell{Arm\\Manipulation} & Multi-Agent & Sound & VR && Engine & \makecell{Interactive\\Editor} \\
        \midrule
        AI2-THOR & $\infty$~\cite{deitke2022procthor} & 3578 && \cmark & \cmark & \cmark & \cmark & \cmark && Unity & \cmark \\
        iGibson 2.0 & 15 & 1217 && \cmark & \cmark & \cmark & \xmark & \cmark && PyBullet & \xmark \\
        Habitat 1.0 & 1000 & \dash && \xmark & \xmark & \xmark & \cmark & \xmark && Magnum & \xmark \\
        Habitat 2.0 & 105 & 92 && \cmark & \cmark & \xmark & \xmark & \xmark && Magnum & \xmark \\
        ThreeDWorld & 15 & 200 && \xmark & \cmark & \xmark & \cmark & \cmark && Unity & \cmark \\
        SAPIEN & 0 & 2346 && \xmark & \cmark & \xmark & \xmark & \xmark && PhysX & \xmark \\
        \bottomrule
    \end{tabular}
    }
    \vspace{0.075in}
    \caption{A comparison table between Embodied AI simulators.}
    \label{tab:comparisons}
\end{table}

Following AI2-THOR's first release in 2017, a number of simulators have been developed, including iGibson 2.0~\cite{li2021igibson}, Habitat 1.0~\cite{savva2019habitat}, Habitat 2.0~\cite{szot2021habitat}, ThreeDWorld~\cite{gan2020threedworld}, and SAPIEN~\cite{xiang2020sapien}. Table~\ref{tab:comparisons} shows a comparison table between the simulators. AI2-THOR is significantly larger in scale than other simulators, while providing first-class support for interaction, and, by leveraging Unity, makes it easy to add new capabilities.

\paragraph{Performance Benchmark.} To benchmark performance, we trained an ObjectNav agent for 1 million steps on a 2-GPU machine. Here, GPU-0 stores and performs updates to the model while GPU-1 renders a batch of parallel instances of the simulator. We obtain a training FPS ranging between 145.5--179.4 (167.7 average). For comparison, we ran the same setup with Habitat 1.0 and obtained a training FPS ranging between 119.7--264.3 (230.5 average). More details are described in Appendix~\ref{sec:performance}.

\section{Conclusion}

We present AI2-THOR, a large-scale interactive simulation platform for Embodied AI. It has been used for experimentation in over 150 publications, spanning a wide variety of tasks and research areas. It is highly customizable, and provides first-class support for many different types of scenes, agent embodiments, actions, and metadata. The capabilities of AI2-THOR are rapidly evolving, and we are excited to support new improvements and use cases to come. For the latest information, please visit our website: \url{https://ai2thor.allenai.org/}.

{
\small

\bibliographystyle{ieee}
\bibliography{egbib}

\begin{thebibliography}{10}\itemsep=-1pt

\bibitem{deitke2020robothor}
Matt Deitke, Winson Han, Alvaro Herrasti, Aniruddha Kembhavi, Eric Kolve,
  Roozbeh Mottaghi, Jordi Salvador, Dustin Schwenk, Eli VanderBilt, Matthew
  Wallingford, Luca Weihs, Mark Yatskar, and Ali Farhadi.
\newblock Robothor: An open simulation-to-real embodied ai platform.
\newblock In {\em CVPR}, 2020.

\bibitem{deitke2022procthor}
Matt Deitke, Eli VanderBilt, Alvaro Herrasti, Luca Weihs, Jordi Salvador, Kiana
  Ehsani, Winson Han, Eric Kolve, Ali Farhadi, Aniruddha Kembhavi, and Roozbeh
  Mottaghi.
\newblock Procthor: Large-scale embodied ai using procedural generation.
\newblock {\em arXiv}, 2022.

\bibitem{du2020learning}
Heming Du, Xin Yu, and Liang Zheng.
\newblock Learning object relation graph and tentative policy for visual
  navigation.
\newblock In {\em ECCV}, 2020.

\bibitem{dwivedi2022navigation}
Kshitij Dwivedi, Gemma Roig, Aniruddha Kembhavi, and Roozbeh Mottaghi.
\newblock What do navigation agents learn about their environment?
\newblock In {\em CVPR}, 2022.

\bibitem{ehsani2021manipulathor}
Kiana Ehsani, Winson Han, Alvaro Herrasti, Eli VanderBilt, Luca Weihs, Eric
  Kolve, Aniruddha Kembhavi, and Roozbeh Mottaghi.
\newblock Manipulathor: A framework for visual object manipulation.
\newblock In {\em CVPR}, 2021.

\bibitem{ehsani2018segan}
Kiana Ehsani, Roozbeh Mottaghi, and Ali Farhadi.
\newblock Segan: Segmenting and generating the invisible.
\newblock In {\em CVPR}, 2018.

\bibitem{gan2020threedworld}
Chuang Gan, Jeremy Schwartz, Seth Alter, Martin Schrimpf, James Traer, Julian
  De~Freitas, Jonas Kubilius, Abhishek Bhandwaldar, Nick Haber, Megumi Sano,
  et~al.
\newblock Threedworld: A platform for interactive multi-modal physical
  simulation.
\newblock In {\em Neural Information Processing Systems Datasets and Benchmarks
  Track (Round 1)}, 2020.

\bibitem{gan2020look}
Chuang Gan, Yiwei Zhang, Jiajun Wu, Boqing Gong, and Joshua~B Tenenbaum.
\newblock Look, listen, and act: Towards audio-visual embodied navigation.
\newblock In {\em ICRA}, 2020.

\bibitem{gao2022dialfred}
Xiaofeng Gao, Qiaozi Gao, Ran Gong, Kaixiang Lin, Govind Thattai, and Gaurav~S
  Sukhatme.
\newblock Dialfred: Dialogue-enabled agents for embodied instruction following.
\newblock {\em IEEE Robotics and Automation Letters}, 2022.

\bibitem{gordon2018iqa}
Daniel Gordon, Aniruddha Kembhavi, Mohammad Rastegari, Joseph Redmon, Dieter
  Fox, and Ali Farhadi.
\newblock Iqa: Visual question answering in interactive environments.
\newblock In {\em CVPR}, 2018.

\bibitem{CordialSync}
Unnat Jain, Luca Weihs, Eric Kolve, Ali Farhadi, Svetlana Lazebnik, Aniruddha
  Kembhavi, and Alexander~G. Schwing.
\newblock A cordial sync: Going beyond marginal policies for multi-agent
  embodied tasks.
\newblock In {\em ECCV}, 2020.

\bibitem{TwoBody}
Unnat Jain, Luca Weihs, Eric Kolve, Mohammad Rastegari, Svetlana Lazebnik, Ali
  Farhadi, Alexander~G. Schwing, and Aniruddha Kembhavi.
\newblock Two body problem: Collaborative visual task completion.
\newblock In {\em CVPR}, 2019.

\bibitem{karamcheti2020learning}
Siddharth Karamcheti, Dorsa Sadigh, and Percy Liang.
\newblock Learning adaptive language interfaces through decomposition.
\newblock {\em arXiv}, 2020.

\bibitem{kemp2022design}
Charles~C Kemp, Aaron Edsinger, Henry~M Clever, and Blaine Matulevich.
\newblock The design of stretch: A compact, lightweight mobile manipulator for
  indoor human environments.
\newblock In {\em ICRA}, 2022.

\bibitem{khandelwal2022simple}
Apoorv Khandelwal, Luca Weihs, Roozbeh Mottaghi, and Aniruddha Kembhavi.
\newblock Simple but effective: Clip embeddings for embodied ai.
\newblock In {\em CVPR}, 2022.

\bibitem{kotar2021contrasting}
Klemen Kotar, Gabriel Ilharco, Ludwig Schmidt, Kiana Ehsani, and Roozbeh
  Mottaghi.
\newblock Contrasting contrastive self-supervised representation learning
  pipelines.
\newblock In {\em ICCV}, 2021.

\bibitem{kotar2022interactron}
Klemen Kotar and Roozbeh Mottaghi.
\newblock Interactron: Embodied adaptive object detection.
\newblock In {\em CVPR}, 2022.

\bibitem{li2021igibson}
Chengshu Li, Fei Xia, Roberto Mart{\'\i}n-Mart{\'\i}n, Michael Lingelbach,
  Sanjana Srivastava, Bokui Shen, Kent~Elliott Vainio, Cem Gokmen, Gokul
  Dharan, Tanish Jain, Andrey Kurenkov, Karen Liu, Hyowon Gweon, Jiajun Wu, Li
  Fei-Fei, and Silvio Savarese.
\newblock igibson 2.0: Object-centric simulation for robot learning of everyday
  household tasks.
\newblock In {\em CoRL}, 2021.

\bibitem{li2021ifr}
Qi Li, Kaichun Mo, Yanchao Yang, Hang Zhao, and Leonidas Guibas.
\newblock Ifr-explore: Learning inter-object functional relationships in 3d
  indoor scenes.
\newblock In {\em ICLR}, 2022.

\bibitem{liu2022multi}
Xinzhu Liu, Di Guo, Huaping Liu, and Fuchun Sun.
\newblock Multi-agent embodied visual semantic navigation with scene prior
  knowledge.
\newblock {\em IEEE Robotics and Automation Letters}, 2022.

\bibitem{lohmann2020learning}
Martin Lohmann, Jordi Salvador, Aniruddha Kembhavi, and Roozbeh Mottaghi.
\newblock Learning about objects by learning to interact with them.
\newblock In {\em NeurIPS}, 2020.

\bibitem{lu2021mgrl}
Yi Lu, Yaran Chen, Dongbin Zhao, and Dong Li.
\newblock Mgrl: Graph neural network based inference in a markov network with
  reinforcement learning for visual navigation.
\newblock {\em Neurocomputing}, 2021.

\bibitem{min2021film}
So~Yeon Min, Devendra~Singh Chaplot, Pradeep Ravikumar, Yonatan Bisk, and
  Ruslan Salakhutdinov.
\newblock Film: Following instructions in language with modular methods.
\newblock In {\em ICLR}, 2022.

\bibitem{murali2019pyrobot}
Adithyavairavan Murali, Tao Chen, Kalyan~Vasudev Alwala, Dhiraj Gandhi, Lerrel
  Pinto, Saurabh Gupta, and Abhinav Gupta.
\newblock Pyrobot: An open-source robotics framework for research and
  benchmarking.
\newblock {\em arXiv}, 2019.

\bibitem{nagarajan2020learning}
Tushar Nagarajan and Kristen Grauman.
\newblock Learning affordance landscapes for interaction exploration in 3d
  environments.
\newblock In {\em NeurIPS}, 2020.

\bibitem{nagarajan2021shaping}
Tushar Nagarajan and Kristen Grauman.
\newblock Shaping embodied agent behavior with activity-context priors from
  egocentric video.
\newblock In {\em NeurIPS}, 2021.

\bibitem{padmakumar2022teach}
Aishwarya Padmakumar, Jesse Thomason, Ayush Shrivastava, Patrick Lange, Anjali
  Narayan-Chen, Spandana Gella, Robinson Piramuthu, Gokhan Tur, and Dilek
  Hakkani-Tur.
\newblock Teach: Task-driven embodied agents that chat.
\newblock In {\em AAAI}, 2022.

\bibitem{pashevich2021episodic}
Alexander Pashevich, Cordelia Schmid, and Chen Sun.
\newblock Episodic transformer for vision-and-language navigation.
\newblock In {\em ICCV}, 2021.

\bibitem{ramakrishnan2021habitat}
Santhosh~Kumar Ramakrishnan, Aaron Gokaslan, Erik Wijmans, Oleksandr Maksymets,
  Alexander Clegg, John~M Turner, Eric Undersander, Wojciech Galuba, Andrew
  Westbury, Angel~X Chang, Manolis Savva, Yili Zhao, and Dhruv Batra.
\newblock Habitat-matterport 3d dataset ({HM}3d): 1000 large-scale 3d
  environments for embodied {AI}.
\newblock In {\em Neural Information Processing Systems Datasets and Benchmarks
  Track (Round 2)}, 2021.

\bibitem{savva2019habitat}
Manolis Savva, Abhishek Kadian, Oleksandr Maksymets, Yili Zhao, Erik Wijmans,
  Bhavana Jain, Julian Straub, Jia Liu, Vladlen Koltun, Jitendra Malik, Devi
  Parikh, and Dhruv Batra.
\newblock Habitat: A platform for embodied ai research.
\newblock In {\em ICCV}, 2019.

\bibitem{shridhar2020alfred}
Mohit Shridhar, Jesse Thomason, Daniel Gordon, Yonatan Bisk, Winson Han,
  Roozbeh Mottaghi, Luke Zettlemoyer, and Dieter Fox.
\newblock Alfred: A benchmark for interpreting grounded instructions for
  everyday tasks.
\newblock In {\em CVPR}, 2020.

\bibitem{szot2021habitat}
Andrew Szot, Alexander Clegg, Eric Undersander, Erik Wijmans, Yili Zhao, John
  Turner, Noah Maestre, Mustafa Mukadam, Devendra~Singh Chaplot, Oleksandr
  Maksymets, Aaron Gokaslan, Vladimir Vondrus, Sameer Dharur, Franziska Meier,
  Wojciech Galuba, Angel~X. Chang, Zsolt Kira, Vladlen Koltun, Jitendra Malik,
  Manolis Savva, and Dhruv Batra.
\newblock Habitat 2.0: Training home assistants to rearrange their habitat.
\newblock In {\em NeurIPS}, 2021.

\bibitem{tan2020multi}
Sinan Tan, Weilai Xiang, Huaping Liu, Di Guo, and Fuchun Sun.
\newblock Multi-agent embodied question answering in interactive environments.
\newblock In {\em ECCV}, 2020.

\bibitem{weihs2021visual}
Luca Weihs, Matt Deitke, Aniruddha Kembhavi, and Roozbeh Mottaghi.
\newblock Visual room rearrangement.
\newblock In {\em CVPR}, 2021.

\bibitem{weihs2019learning}
Luca Weihs, Aniruddha Kembhavi, Kiana Ehsani, Sarah~M Pratt, Winson Han, Alvaro
  Herrasti, Eric Kolve, Dustin Schwenk, Roozbeh Mottaghi, and Ali Farhadi.
\newblock Learning generalizable visual representations via interactive
  gameplay.
\newblock In {\em ICLR}, 2021.

\bibitem{weihs2020allenact}
Luca Weihs, Jordi Salvador, Klemen Kotar, Unnat Jain, Kuo{-}Hao Zeng, Roozbeh
  Mottaghi, and Aniruddha Kembhavi.
\newblock Allenact: {A} framework for embodied {AI} research.
\newblock {\em arXiv}, 2020.

\bibitem{wortsman2019learning}
Mitchell Wortsman, Kiana Ehsani, Mohammad Rastegari, Ali Farhadi, and Roozbeh
  Mottaghi.
\newblock Learning to learn how to learn: Self-adaptive visual navigation using
  meta-learning.
\newblock In {\em CVPR}, 2019.

\bibitem{wu2021communicative}
Qi Wu, Cheng-Ju Wu, Yixin Zhu, and Jungseock Joo.
\newblock Communicative learning with natural gestures for embodied navigation
  agents with human-in-the-scene.
\newblock In {\em IROS}, 2021.

\bibitem{xiang2020sapien}
Fanbo Xiang, Yuzhe Qin, Kaichun Mo, Yikuan Xia, Hao Zhu, Fangchen Liu, Minghua
  Liu, Hanxiao Jiang, Yifu Yuan, He Wang, Li Yi, Angel~X. Chang, Leonidas~J.
  Guibas, and Hao Su.
\newblock Sapien: A simulated part-based interactive environment.
\newblock In {\em CVPR}, 2020.

\bibitem{yang2018visual}
Wei Yang, Xiaolong Wang, Ali Farhadi, Abhinav Gupta, and Roozbeh Mottaghi.
\newblock Visual semantic navigation using scene priors.
\newblock In {\em ICLR}, 2019.

\bibitem{zellers2021piglet}
Rowan Zellers, Ari Holtzman, Matthew~E. Peters, Roozbeh Mottaghi, Aniruddha
  Kembhavi, Ali Farhadi, and Yejin Choi.
\newblock Piglet: Language grounding through neuro-symbolic interaction in a 3d
  world.
\newblock In {\em ACL}, 2021.

\bibitem{zeng2020visual}
Kuo-Hao Zeng, Roozbeh Mottaghi, Luca Weihs, and Ali Farhadi.
\newblock Visual reaction: Learning to play catch with your drone.
\newblock In {\em CVPR}, 2020.

\bibitem{zhao2021luminous}
Yizhou Zhao, Kaixiang Lin, Zhiwei Jia, Qiaozi Gao, Govind Thattai, Jesse
  Thomason, and Gaurav~S Sukhatme.
\newblock Luminous: Indoor scene generation for embodied ai challenges.
\newblock {\em arXiv}, 2021.

\bibitem{zheng2022towards}
Kaiyu Zheng, Rohan Chitnis, Yoonchang Sung, George Konidaris, and Stefanie
  Tellex.
\newblock Towards optimal correlational object search.
\newblock In {\em ICRA}, 2022.

\bibitem{zhu2017target}
Yuke Zhu, Roozbeh Mottaghi, Eric Kolve, Joseph~J Lim, Abhinav Gupta, Li
  Fei-Fei, and Ali Farhadi.
\newblock Target-driven visual navigation in indoor scenes using deep
  reinforcement learning.
\newblock In {\em ICRA}, 2017.

\end{thebibliography}

}

\appendix

\section{Contributions}

\paragraph{Eric Kolve} was the lead engineer and built the API that connects Python and Unity, setup the infrastructure for maintenance and development, heavily optimized AI2-THOR to run faster, added support for headless rendering, contributed to the Unity backend, and contributed to RoboTHOR, ProcTHOR, and ManipulaTHOR.

\paragraph{Roozbeh Mottaghi} managed the AI2-THOR project and its constituents and made decisions about the technical and artistic features of the framework and set priorities for the team.

\paragraph{Winson Han} contributed to the Unity backend logic for features and functionality across AI2-THOR; oversaw the design and functionality of the agents; led the development of logic to support physics-based object interactions, state changes, visibility, repositioning, and the annotation pipeline; set up default object placement in scenes; contributed to the documentation; managed community feature requests and issues; and created many promotional graphics.

\paragraph{Eli VanderBilt} built all of the 3D scenes for iTHOR, RoboTHOR, ArchitecTHOR; created thousands of interactive assets; modeled the agents; and designed and implemented various features, including arm-based manipulation.

\paragraph{Luca Weihs} contributed to the AI2-THOR frontend and backend through the creation of new actions, tests, and processes; led the development of the AllenAct framework, a library used to train agents on AI2-THOR and Embodied AI tasks~\cite{weihs2020allenact}.

\paragraph{Alvaro Herrasti} developed features and infrastructure for the Unity backend and Python API; graphics and shader work; built the WebGL infrastructure and demo integration; built the continuous action physics system for arm-based agents; led the Unity development of ProcTHOR; and contributed to RoboTHOR and ManipulaTHOR.

 \paragraph{Matt Deitke} led the development of ProcTHOR; built the AI2-THOR website, demo, and wrote documentation; contributed to building RoboTHOR; built infrastructure to make AI2-THOR more accessible; contributed to the Unity backend and Python API; and wrote the revised paper.

\paragraph{Kiana Ehsani} led the ManipulaTHOR project and the direction of adding arm-based manipulation with the StretchRE1 and ManipulaTHOR agents.

\paragraph{Daniel Gordon} developed some planning and rendering features for the early versions of AI2-THOR. 

\paragraph{Yuke Zhu} created the very first version of AI2-THOR (mentioned in \cite{zhu2017target}) with the help of EK and RM.

\paragraph{Aniruddha Kembhavi} was involved in decision making for various features of ProcTHOR, ManipulaTHOR, ArchitecTHOR, and RoboTHOR.

\paragraph{Abhinav Gupta} provided advice and guidance throughout the course of the project.
\paragraph{Ali Farhadi} provided advice and guidance throughout the course of the project.

\section{Performance Comparison}
\label{sec:performance}

Comparing performance between Embodied AI simulators is a surprisingly difficult question for many reasons:
\begin{enumerate}[leftmargin=0.25in]
    \item Different simulators support different agents, each with their own action spaces and capabilities, with little standardization across simulators. AI2-THOR supports many different types of agents, including the ManipulaTHOR, Abstract, and LoCoBot agents. The ManipulaTHOR agent is often slower to simulate than a navigation-only LoCoBot agent as it is more complex to physically model a 6 DoF arm as it interacts with objects. This is made even more complex when noting that random action sampling, the simplest policy with which to benchmark, is a poor profiling strategy as some actions are only computationally expensive in rare, but important, settings; for instance, computing arm movements is most expensive when the arm is interacting with many objects, these interactions are rare when randomly sampling but we'd expect them to dominate when using a well-trained agent.
    \item Some simulators are relatively slow when run on a single process but can be easily parallelized with many processes running on a single GPU, \eg AI2-THOR. Thus single-process simulation speeds may be highly deceptive as they do not capture the ease of scalability.
    \item When training agents via reinforcement learning, there are a large number of factors that bottleneck training speed and so the value of raw simulator speed is substantially reduced. These factors include:
    \begin{enumerate}
        \item Model forward pass when computing agent rollouts.
        \item Model backward pass when computing gradients for RL losses.
        \item Environment resets - for many simulators (e.g. AI2-THOR, Habitat, iGibson) it is orders of magnitude more expensive to change a scene than it is to take a single agent step. This can be extremely problematic when using synchronous RL algorithms as all simulators will need to wait for a single simulator when that simulator is resetting. When training this means that, in practice, important "tricks" are employed to ensure that scene changes are infrequent or synchronized, without these tricks, performance may be dramatically lower.
    \end{enumerate}
\end{enumerate}

To attempt to control for the above factors, we set up two profiling experiments, one in Habitat with HM3D~\cite{ramakrishnan2021habitat} and one using ProcTHOR-10K, where we:
\begin{itemize}[leftmargin=0.25in]
    \item Use a 2-GPU machine (GeForce RTX 2080 GPUs) where GPU-0 is reserved for the agent's actor-critic policy network and GPU-1 is reserved for simulator instances.
    \item Train agents for the ObjectNav task (using the same LoCoBot agent with the same action space).
    \item For both agents, use the same actor-critic policy network, the same used in the ProcTHOR paper~\cite{deitke2022procthor}.
    \item Remove the "End" action so that agents always take the maximum 500 steps, this minimizes dependence on the learned policy.
    \item Use a rollout length of 128 with the same set of training hyperparameters across both models.
    \item Use a total of 28 parallel simulator processes, this approximately saturates GPU-1 memory. We found that Habitat instances used slightly less GPU memory than ProcTHOR instances and so we could likely increase the number instances for Habitat slightly, but we kept these equal for more direct comparison.
    \item Use a scene update "trick" which forces all simulators to advance to the next scene in a synchronous fashion after every 10 rollouts (e.g. after every 10 x 128 x 28 = 35,840 total steps across all simulators).
\end{itemize}

\end{document}